\documentclass[lettersize,journal]{IEEEtran}
\usepackage{amsmath,amssymb,amsfonts}
\usepackage{algorithm}
\usepackage{algorithmic}
\usepackage{array}
\usepackage[caption=true,font=normalsize,labelfont=sf,textfont=sf]{subfig}
\usepackage{textcomp}
\usepackage{stfloats}
\usepackage{url}
\usepackage{cite}
\usepackage{verbatim}
\usepackage{graphicx}
\usepackage[T1]{fontenc}

\usepackage{epstopdf}
\usepackage{multirow}
\usepackage{color}
\usepackage{pifont}

\renewcommand{\&}{and}
\usepackage{hyperref}
\hypersetup{colorlinks=true,linkcolor=blue,citecolor=blue,urlcolor=black}
\usepackage{balance}
\usepackage{cleveref}

\begin{document}
\title{Robust Fuzzy local k-plane clustering with mixture distance of hinge loss and $L_1$ norm}
\author{Junjun Huang, Xiliang Lu, Xuelin Xie, and Jerry Zhijian Yang

\thanks{This work was supported in part by the National Key Research and Development Program of China under Grant 2023YFA1000103, in part by the National Natural Science Foundation of China under Grants 12371424, 12125103, and U24A2002, in part by the Natural Science Foundation of Hubei Province, China under Grant 2024AFE006, and in part by the Fundamental Research Funds for the Central Universities, China. \textit{(Corresponding authors: Xiliang Lu.)} }
\and
\thanks{The authors are listed in alphabetical order by surname. J. Huang, X. Lu, X. Xie, and J. Z. Yang contributed equally to this work.}
\and
\thanks{Junjun Huang is with the China Electric Power Research Institute, Wuhan 430074, China, and also with the Key Laboratory of Measurement and Test of High Voltage and Heavy Current, State Administration for Market Regulation, Wuhan 430074, China (email: huangjunjun@epri.sgcc.com.cn).}
\and
\thanks{Xiliang Lu is with the School of Mathematics and Statistics, Wuhan University, Wuhan 430072, China, the National Center for Applied Mathematics in Hubei, Wuhan University, Wuhan 430072, China, and also with the Hubei Key Laboratory of Computational Science, Wuhan University, Wuhan 430072, China (e-mail: xllv.math@whu.edu.cn).}
\and
\thanks{Xuelin Xie is with the School of Mathematics and Statistics, Wuhan University, Wuhan 430072, China (e-mail: xl.xie@whu.edu.cn).}
\and
\thanks{Jerry Zhijian Yang is with the National Center for Applied Mathematics in Hubei, Wuhan University, Wuhan 430072, China, the Wuhan Institute for Math \& AI, Wuhan University, Wuhan 430072, China, the School of Mathematics and Statistics, Wuhan University, Wuhan 430072, China, and the Hubei Key Laboratory of Computational Science, Wuhan University, Wuhan 430072, China (e-mail: zjyang.math@whu.edu.cn).}
}
\markboth{T\lowercase{his} M\lowercase{anuscript} \lowercase{is accepted by} \textit{IEEE TKDE, 2025.}}%
{How to Use the IEEEtran \LaTeX \ Templates}

\maketitle

\begin{abstract}

K-plane clustering (KPC), hyperplane clustering, and mixture regression all essentially fall within the same class of problems. This problem can be conceptualized as clustering in relatively high-dimensional K subspaces or K linear manifolds. Traditional KPC or fuzzy KPC models demonstrate a pronounced susceptibility to outliers, as they presuppose that the projection distance between data points and the plane normal vector adheres to the $L_2$ distance. Meanwhile, the assumption of infinitely extending clusters adversely affects clustering performance. To solve these problems, this paper proposed a new robust fuzzy local k-plane clustering (RFLkPC) method that combines the mixture distance of hinge loss and $L_1$ norm. The RFLkPC model assumes that each plane cluster is bounded to a finite area, which can flexibly and robustly handle plane clustering tasks with outliers or not. The corresponding model and optimization algorithms of RFLkPC were provided. Compared to other related models on this topic, a large number of experiments verify the efficiency of RFLkPC on simulated data and real data. The source code for the proposed RFLkPC method is publicly available at \url{https://github.com/xuelin-xie/RFLkPC}.

\end{abstract}


\section{INTRODUCTION}
\IEEEPARstart{P}{lane} clustering assumes that the manifold structure of each subcluster is a hyperplane, while traditional clustering considers each subcluster as a sphere-type cluster, such as the popular K-means \cite{MacQuuen1967}, fuzzy c-means (FCM) \cite{Bezdek1994}. Therefore, K-plane clustering can be considered a class of clustering tasks with special structures. Despite the particular cluster structure of each group, its applications are extremely rich, involving economics, statistics, computer vision, pattern recognition, image segmentation, etc. However, due to the differences in the fields of the researchers, the name of this topic varies, which affects the communication and development of this research to some extent.

For example, when Quandt and Goldfeld \cite{Goldfeld1973} studied economic phenomena, they found that the samples' regression relationship changed after a certain time, and they called it "switching regression" (SR). In statistics, to analyze the heterogeneous growth of animals or plants, it is assumed that data may come from a mixture model of multiple regression \cite{huang2017regression}.
Therefore, it was also called "mixture regression model" (MRM), such as \cite{song2014,shan2021}, or named "clusterwise linear regression" (CLR) in \cite{spath1979,schlittgen2011}. In computer vision or subspace clustering, some applications can often be reduced to hyperplane clustering (HC) \cite{tsakiris2017,ding2021dual}, such as motion segmentation \cite{tron2007,chen2009}, identification of hybrid linear systems \cite{bako2011,blavzivc2020}. In addition, followed by classical fuzzy c-regression model (FCRM), fuzzy c-varieties or regression clustering, a lot of related research can be found in \cite{yang2008,leski2015,nie2020,blavzivc2019,zhao2021,hu2022,leski2023}. These different titles have similar themes, and they can all be regarded as clustering tasks with multiple clusters of plane structures. It should be noted that, based on SR, MRM, CLR, or FCRM, it is assumed that prior knowledge exists regarding which attributes of the data are explanatory variables and which are the response variables. Compared to plane clustering or hyperplane clustering, this assumption restricts the broader application of mixed regression models. Therefore, this paper will focus on the extension of plane clustering.

Plane-based clustering models, such as KPC \cite{bradley2000k}, do not need to know what the response variable of the sample is, and only focus on the estimation of K-plane normal vectors by solving a series of matrix eigenvalue problems. Unfortunately, due to the sensitivity of clustering and plane estimation to parameter initialization and outliers, the combination of the two is particularly prone to get stuck in local optima in plane clustering tasks, which limits its widespread application. Therefore, researchers have made numerous improvements and extensions. For example, Shao et al. improved KPC to k-proximal plane clustering (KPPC) \cite{shao2013} by considering the within-cluster distance and the between-cluster distance at the same time. Yang et al. made full use of the advantages of local information and inter-cluster distance, and proposed local k-proximal plane clustering (LkPPC) \cite{yang2015local}. Another type of models that considered both intra-cluster and inter-cluster distances are twin support vector clustering models (TWSVC) \cite{wang2015twin}, and related robust improvements \cite{ye2017l1,bai2019clustering,richhariya2020,tanveer2021sparse}.  K-flats \cite{tseng2000} and localized k-flats (LkF) \cite{wang2011} extended the clustering of plane clusters to the clustering of flats.

In the aforementioned plane-based clustering algorithms, all of them are hard clustering methods, which means that each sample is uniquely assigned to one category. In many real-world applications or overlapping cases, a sample may belong to more than one cluster. Therefore, Zhu et al. suggested fuzzy k-plane clustering (FkPC) \cite{zhu2008} by utilizing fuzzy set theory to represent the partition matrix in contrast to hard clustering. Recently, Gu et al. extended FkPC to style clustering and proposed the fuzzy style k-plane clustering (FSkPC) \cite{gu2020fuzzy} method. Puneet Kumar et al. \cite{kumar2022} successfully applied FkPC to the segmentation of a human brain medical image by incorporating the local spatial information of the image. These models reflect the flexibility and powerful capabilities of fuzzy k-plane clustering applications.

{To overcome the the {{influence}} of outliers in plane-based hard clustering, in addition to the methods mentioned earlier \cite{ye2017l1} with $L_1$ norm and based on pinball loss (proposed by Huang \cite{huang2013support}), Qi et al. \cite{QI2023locally} recently introduced the locally finite capped $\ell_{2,1}$-norm into plane clustering to avoid the infinite extension of clusters and sensitivity to outliers. However, within the framework of fuzzy plane clustering, to the best of our knowledge, there is relatively limited research on robust distance design and constraints on the boundedness of plane clusters for FkPC \cite{zhu2008}.}

{{Motivated by the limitations of existing approaches, we propose a robust fuzzy local k-plane clustering (RFLkPC) model that simultaneously tackles two key challenges: sensitivity to outliers and unbounded cluster extensions. Drawing inspiration from the local boundedness constraint in the LkF model~\cite{wang2011}, RFLkPC incorporates a squared distance regularization term that constrains points to remain close to their respective cluster centers, effectively preventing excessive plane expansion. In addition, we introduce a hybrid distance function that blends the $L_2$ and $L_1$ norm, inspired by the elastic-net regularization \cite{zou2005regularization}. This design enhances the model’s ability to resist heavy-tailed noise and large residuals-common challenges in real-world datasets.}}

\begin{figure}[!h]
	\centering
	\begin{minipage}{0.45\linewidth}
		\centering
		\includegraphics[width=0.9\linewidth]{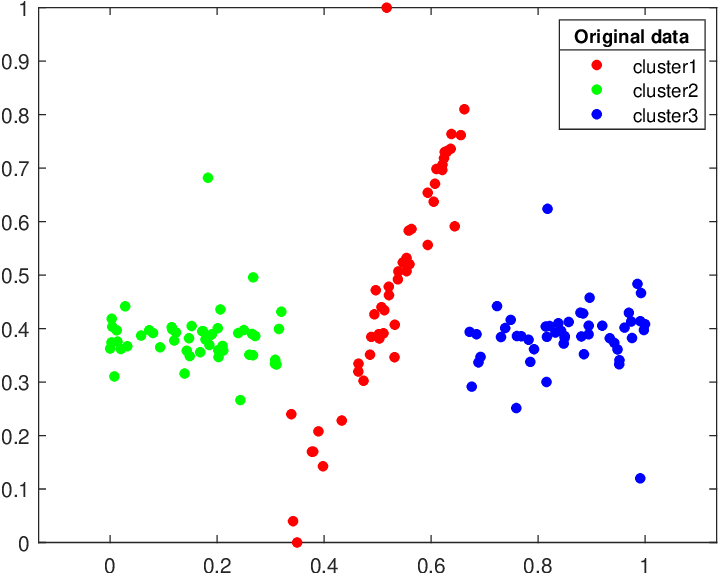}
	\end{minipage}
	\begin{minipage}{0.45\linewidth}
		\centering
		\includegraphics[width=0.9\linewidth]{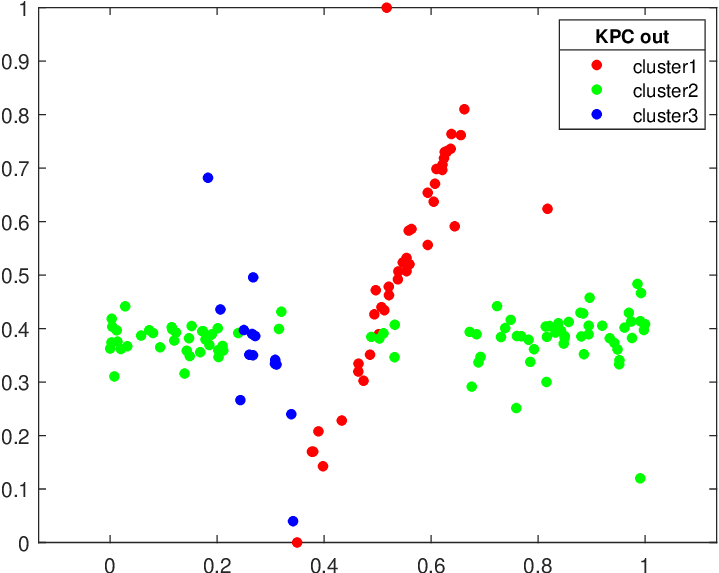}
	\end{minipage}
	\vspace{3mm}
	\begin{minipage}{0.45\linewidth}
		\centering
		\includegraphics[width=0.9\linewidth]{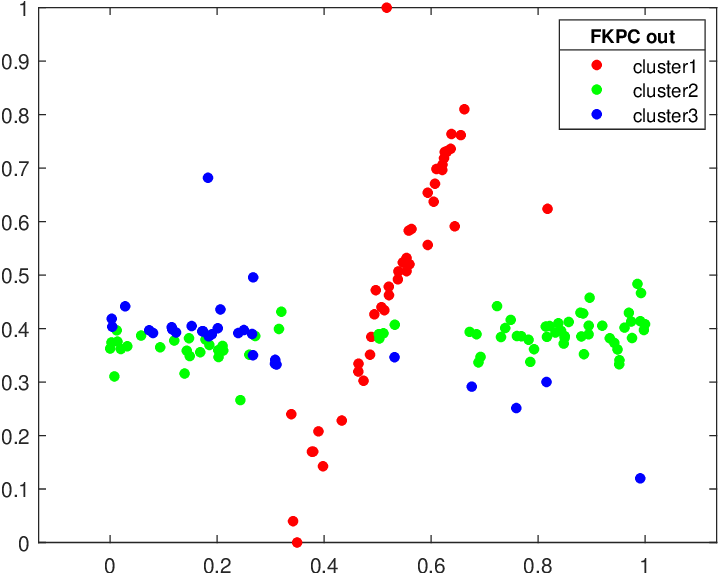}
	\end{minipage}
	\begin{minipage}{0.45\linewidth}
		\centering
		\includegraphics[width=0.9\linewidth]{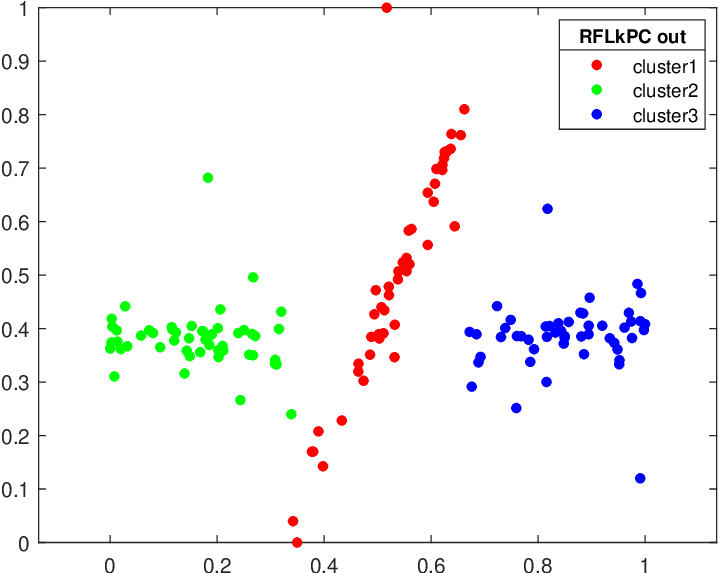}
	\end{minipage}
    \caption{{A toy example with three linear-structure clusters contaminated by heavy-tailed noise. Clustering results are shown for KPC (top-right), FKPC (bottom-left), and the proposed RFLkPC (bottom-right), illustrating the effectiveness of RFLkPC in separating clusters with similar directions under noise.}}
    \label{fig-toy}
\end{figure}

A synthetic example (Fig.~\ref{fig-toy}) illustrates the effectiveness of the proposed model on three linear-structured clusters corrupted by heavy-tailed noise, with clusters 2 and 3 exhibiting similar orientations. Compared to classical methods such as KPC \cite{bradley2000k} and FkPC \cite{zhu2008}, our model successfully separates clusters with overlapping directions, demonstrating the benefit of incorporating both boundedness and robustness into the fuzzy clustering framework.

The main contributions of this paper are summarized as follows:
\begin{enumerate}
\item We propose a robust fuzzy local k-plane clustering model that jointly leverages local boundedness and a hybrid distance metric to enhance resilience to noise and outliers. Meanwhile, a corresponding optimization algorithm is developed to solve the RFLkPC model.
\item Considering the interdisciplinary nature of this topic, we survey relevant work on plane-based clustering in different research fields as much as possible, hoping to facilitate better communication and development on this topic.
\item Extensive experiments on both artificial and real-world datasets show that the proposed RFLkPC performs better than previous related methods.
\end{enumerate}

The rest of this paper proceeds as follows. In Section \ref{sec2}, we will provide some details of models on related research, including KPC models, FkPC, MRMs, FCRMs, and other approaches that target improving the accuracy of plane clustering. In Section \ref{sec3}, we describe the RFLkPC model and the corresponding optimization algorithm. We compare the experimental results of our RFLkPC algorithm with those of other algorithms to demonstrate its efficiency in Section \ref{sec4}. Finally, we present conclusions and directions for future research in Section \ref{sec5}.

\section{RELATED WORK}  \label{sec2}
In this section, we specifically introduce some related plane clustering models mentioned above, including plane clustering models, mixture regression models, fuzzy c-regression models, and others. We have included as much as possible the classic methods or perspectives of different research fields on this topic, in order to achieve better communication and in-depth research on this topic. Additionally, we further review some robust models based on the estimation of multiple regression relationships, and high-dimensional k-subspace clustering methods such as DPCP-KSS \cite{ding2021dual} of hyperplane clustering.

\subsection{Plane clustering models}
Consider a set of data points $x_i$ $(i=1,\dots,N)$ in the $D$-dimensional space represented by the data matrix $A=(x_1,\dots,x_N)\in \mathbb{R}^{D\times N}$, KPC \cite{bradley2000k} clusters the data $A$ into $K$ clusters $C_k$ $(k=1,\dots,K)$ such that the data points of each cluster are grouped around the plane cluster defined as follows:
\begin{align}\label{KPC}
  \min_{\mathbf{v_k},b_k} & \sum_{k=1}^{K}\sum_{A_k\in C_k}\|A_k^{T}\mathbf{v_k}+b_k\mathbf{e_k}\|_{2}^2\\
  \nonumber & s.t.  \|\mathbf{v_k}\|_{2} =1,(\forall k=1,\dots,K),
\end{align}
where $A_k\in \mathbb{R}^{D\times n_k}$ is the data points of the kth cluster $C_k$, $\mathbf{e_k}$ is a vector of all ones of appropriate dimension, $\mathbf{v_k}\in \mathbb{R}^{D\times 1}$ is the unit normal vector corresponding to each plane cluster, $b_k$ is a scalar, and the parameter pair $(\mathbf{v_k},b_{k})$ determines a plane cluster. The solution of model \eqref{KPC} can be obtained by iteratively solving eigenvalue problems \cite{bradley2000k}.

Compared with KPC, KPPC \cite{shao2013} not only requests the objective data points closest to the corresponding plane, but also lets the other data points be far away from this center plane, which leads to solving the following model:
\begin{align}\label{KPPC}
  \min_{\mathbf{v_k},b_k} & \sum_{k=1}^{K}\sum_{A_k\in C_k}\left(\|A_k^{T}\mathbf{v_k}+b_k\mathbf{e_k}\|_{2}^2 - c\|B_k^{T}\mathbf{v_k}+b_k\mathbf{\overline{e_k}}\|_2^2\right)\\
  \nonumber & s.t. \|\mathbf{v_k}\|_{2}  =1,(\forall k=1,\dots,K),
\end{align}
where $A_k$, $\mathbf{e_k}$, $\mathbf{v_k}$, $b_k$ have the same meaning as \eqref{KPC}, $B_k\in \mathbb{R}^{D\times (N-n_k)}\subseteq A$ is the data points not belonging to the kth plane cluster, $\mathbf{\overline{e_k}}$ is a vector of all ones of appropriate dimension, $c>0$ is a weight parameter to balance the proximal term and the aloof term.

By introducing local information, Yang et al. \cite{yang2015local} gave the local kPPC model (LkPPC):
\begin{align}\label{LkPPC}
  \nonumber \min_{\mathbf{v_k},b_k} & \sum_{k=1}^{K}\sum_{A_k\in C_k}(\|A_k^{T}\mathbf{v_k}+b_k\mathbf{e_k}\|_{2}^2 +c_1\|A_k-\mathbf{\mu_k}\mathbf{e_k^T}\|_2^{2} - \\
    & \quad \quad \quad \quad c_2\|B_k^{T}\mathbf{v_k}+b_k\mathbf{\overline{e_k}}\|_2^2)\\
  \nonumber s.t.  &\|\mathbf{v_k}\|_{2}  =1,(\forall k=1,\dots,K),
\end{align}
where the model weights $1 > c_1 > 0$, $c_2 > 0$. $c_1$ controls the localized degree between data and the plane center. $c_2$ is the balance between inter-cluster distance and intra-cluster distance. The difference between the optimization of LkPPC and KPC and KPPC is mainly in the iterative format of the eigenvector solution.

Moreover, based on fuzzy set theory, Zhu et al. proposed the fuzzy k-plane clustering (FkPC) \cite{zhu2008} model:
\begin{align}\label{FkPC}
  \min_{u_{ik},\mathbf{v_k},b_k}  &\sum_{k=1}^{K}\sum_{i=1}^{N} u_{ik}^{m}(x_i^{T}\mathbf{v_k}+b_k)^2\\
  \nonumber & s.t.  \|\mathbf{v_k}\|_{2} =1,\sum_{k=1}^{K}u_{ik}=1,1\leq i\leq N,1\leq k\leq K,
\end{align}
where $u_{ik}\in[0,1]$ is the fuzzy membership value that reflects the degree of the ith sample $x_i$ belonging to the kth cluster, and $m>1$ is the fuzzification factor in fuzzy clustering.

The above optimization \eqref{FkPC} can be reduced to eigenvalue problems using the Lagrange multiplier method, and the kth cluster's normal vector $\mathbf{v_k}$ satisfies the following equation:
\begin{equation}\label{Eqeig}
  W_{k}\mathbf{v_k}=\xi_k\mathbf{v_k},
\end{equation}
where $\mathbf{v_k}$ and $\xi_k$ are the eigenvector and eigenvalue of the kth cluster. The characteristic matrix $W_{k}\in \mathbb{R}^{D\times D}$ of the kth cluster is expressed as:
\begin{equation}\label{Eqeig}
  W_{k}=\sum_{i=1}^{N}u_{ik}^{m}x_{i}x_{i}^T - \frac{(\sum_{i=1}^{N}u_{ik}^{m}x_i)(\sum_{i=1}^{N}u_{ik}^{m}x_{i}^T)}{\sum_{i=1}^{N}u_{ik}^{m}}.
\end{equation}

The plane parameter $\mathbf{v_k}$ is an eigenvector corresponding to the smallest eigenvalue of $W_{k}$. Based on $\mathbf{v_k}$, $b_k$ can be obtained by:
\begin{equation}\label{bk_eq}
  b_{k}= \frac{-\sum_{i=1}^{N}u_{ik}^{m}x_i^T\mathbf{v_k}}{\sum_{i=1}^{N}u_{ik}^{m}}.
\end{equation}

Using the values $\mathbf{v_k}$ and $b_k$, the membership value $u_{ik}$ can be calculated by:
\begin{equation}\label{bk_eq}
  u_{ik}= \left(\frac{(x_i^{T}\mathbf{v_k}+b_k)^2}{\sum_{c=1}^{K}(x_i^{T}\mathbf{v_c}+b_c)^2}\right)^{\frac{-1}{m-1}}.
\end{equation}

The FkPC method alternatively computes the cluster plane prototype and fuzzy membership matrix until the terminating condition is satisfied. Unlike all the models above, LkF \cite{wang2011} generalizes k-plane clustering to flat clustering with local information as follows:
\begin{align}\label{LkF}
  \min_{u_{ik},\mathbf{V_k},\mathbf{\mu_k}}  &\sum_{k=1}^{K}\sum_{i=1}^{N} u_{ik}\left(\|\mathbf{V_k}^T(x_i-\mathbf{\mu_k})\|_2^2+\lambda\|x_i-\mathbf{\mu_k}\|_2^2\right)\\
  \nonumber & s.t.  \mathbf{V_k}^T \mathbf{V_k} =I,\sum_{k=1}^{K}u_{ik}=1,u_{ik}\in\{0,1\},
\end{align}
where $\mathbf{V_k}\in \mathbb{R}^{D\times (D-d)}$ denotes the normal bases of the kth $d$-dimensional ($0<d<D$) linear flat, $\mathbf{\mu_k}\in \mathbb{R}^{D\times 1}$ is the center of each plane cluster, $\lambda$ $(0<\lambda<1)$ controls the data points being close to the center of the kth plane cluster. In practical applications, LkF clusters $K$ distinct flats by training $M$ $(M>K)$ localized linear models and combining spectral clustering methods. Note that when $d=D-1$, it degenerates into hard plane clustering, and its iterative optimization is also attributed to the problem of eigenvalue solving.

LkF belongs to the hard clustering method and cannot be well applied to situations where the sample distribution has overlapping areas. At the same time, when $d=D-1$, it degenerates into plane clustering, and the estimated normal basis $V_k$ is sensitive to outliers due to its $L_2$ distance. Hence, this motivates us to make corresponding improvements to LkF and further provide our model RFLkPC below.

\subsection{Mixture regression models}
Mixture regression models (MRMs) were introduced by Hosmer \cite{hosmer1978} in econometrics, which can be an extension of finite mixture models \cite{Jain2002,mclachlan2019}. In cluster analysis, the idea of finite mixture models is to assume that the samples are from a mixture of different density distributions, with each component density being associated with a cluster.

MRMs can be stated as follows. Let $Z$ be a latent class variable with
$P(Z_{i}=k|X_i)=\pi_{k}$ for $k =1,2,\dots,K$, where $X_i=(\tilde{x}_i;1)$ is a $D$-dimensional vector ($\tilde{x}_i\in \mathbb{R}^{(D-1)\times 1}$ denotes the sample i's explanation variable). Given $Z_{i}= k$, suppose that the response $y_{i}$ depends on explanation variable $X_i$ in a linear regression way:
\begin{equation}\label{}
  y_{i}=\beta_{k}^{T}X_{i}+\epsilon_{ik},
\end{equation}
where $\beta_{k}\in \mathbb{R}^D$ is the kth plane's regression coefficients, and the error term $\epsilon_{ik}\sim N(0,\sigma_{k}^{2})$. Then the conditional density of $y$ given $X$ can be written as:
\begin{equation}\label{MRM1}
  f(y|X)=\sum_{k=1}^{K}\pi_{k}\phi(y;\beta_{k}^{T}X,\sigma_{k}^{2}),
\end{equation}
and the log-likelihood function for observations ${(X_{1}, y_{1}),\dots,(X_N{}, y_{N})}$ is:
\begin{equation}\label{MRM2}
  \sum_{i=1}^{N}\log\left[\sum_{k=1}^{K}\pi_{k}\phi(y_i;\beta_{k}^{T}X_i,\sigma_{k}^{2})\right],
\end{equation}
where $\pi_{k}\in (0,1)$ denotes the probability for each component $k$ with $\sum_{k=1}^{K}\pi_{k}=1$, $\phi(\bullet;\mu,\sigma^{2})$ is the density function of $N(\mu,\sigma^{2})$. The unknown parameters $(\pi_{k},\beta_{k},\sigma_{k})$ in the model \eqref{MRM1} can be estimated by the maximum likelihood estimator (MLE), which usually maximizes \eqref{MRM2} based on the EM \cite{dempster1977} algorithm.

The traditional estimation of mixture regression models is based on the assumption of normality of component errors and thus is sensitive to outliers. In order to increase its robustness, one approach is to change the assumed distribution of the error term. For example, Song et al. \cite{song2014} used the Laplace distribution instead of the original Gaussian function $\phi(\bullet;\mu,\sigma^{2})$. Yao et al. \cite{yao2014} based on the t-distribution of mixture regression, and \cite{chamroukhi2016} based on the skew-normal distribution for improving classical MRMs. Another approach is to modify the E step or M step in the EM algorithm to achieve robustness. For instance, Bai et al. \cite{bai2012} proposed a modification of the EM algorithm for normal mixtures, by replacing the least squares criterion in the M step with a robust criterion. Recently, a unified approach to robustifying MRMs has been proposed by Naderi et al. \cite{Naderi2023}. A more detailed survey of robust mixture regression can be found in \cite{Li2018,yu2020selective}.

\subsection{Fuzzy c-regression models}
Hathaway and Bezdek \cite{hathaway1993} extended the FCM model to regression clustering called  fuzzy c-regression model (FCRM) as follows:
\begin{align}\label{FCRM}
  \min_{\beta_{k},u_{ik}} &\sum_{k=1}^{K}\sum_{i=1}^{N}u_{ik}^{m}(y_{i}-\beta_{k}^{T}X_{i})^2\\
  \nonumber & s.t.  \sum_{k=1}^{K}u_{ik}=1,u_{ik}\in[0,1],
\end{align}
where $X_{i}$, $y_i$ and $\beta_{k}$ are the same as MRM in the previous section, and $u_{ik}$ denotes the membership value. It alternately optimizes $\beta_{k}$ and $u_{ik}$ according to the following formulas:
\begin{equation}\label{Bk}
    \beta_{k} = (XU_{k}X^{T})^{-1}(XU_{k}Y), \forall k=1,\dots,K,
\end{equation}
\begin{equation}\label{UikFCR}
  u_{ik}= \left(\frac{(y_{i}-\beta_{k}^{T}X_{i})^2}{\sum_{c=1}^{K}(y_{i}-\beta_{c}^{T}X_{i})^2}\right)^{\frac{-1}{m-1}},
\end{equation}
where $X=(X_{i})\in \mathbb{R}^{D\times N}$, $Y=(y_i)\in \mathbb{R}^{N\times 1}$, $U_{k}=\text{diag}(u_{ik})\in \mathbb{R}^{N\times N}$ is a diagonal matrix whose elements are $u_{ik}$.

Due to the sensitivity of both regression and clustering to outliers, Leski and Kotas \cite{leski2015} proposed the fuzzy c-ordered-regressions (FCOR) by combining Huber's M-estimators and Yager's OWA operators to continuously enhance robustness in rejecting outliers,and recent work of fuzzy double ordered C-Regression model (FDOCRM \cite{leski2023}). Yang et al. \cite{yang2008} embedded the cluster core concept into switching regressions and created an FCR$\alpha$ model. Referring to possibilistic c-means (PCM) \cite{krishnapuram1993,Zhang2004}, Kung et al. \cite{kung2013} proposed the possibilistic c-regressions (PCR) clustering algorithm, and stepwise possibilistic c-regressions (SPCR) were introduced by Chang et al. \cite{chang2016}. Based on the distinguished feature of the harmonic average, Zhao et al. \cite{zhao2018} provided fuzzy weighted c-harmonic regressions (FWCHR), in which a dynamic-like weight term is introduced to enhance robustness.

MRMs and FCRMs both need to know which sample attribute is the response variable and estimate the K regression relationships, which is the biggest difference between them and the plane-based clustering method.

\subsection{K subspace clustering for Hyperplane Clustering}
Subspace clustering \cite{vidal2011subspace} assumes data points are drawn from a union of subspaces, and the goal is to estimate the subspaces and cluster the data points according to their partition. Hyperplane clustering (HC) can be viewed as a high-dimensional subspace clustering task in which the dimension of each subspace is $(D-1)$ with respect to D-dimensional data.

There are many mainstream HC methods. For example, Random Sampling and Consensus (RANSAC) \cite{cantzler1981}, Spectral Curvature Clustering (SCC) \cite{chen2009}, Algebraic Subspace Clustering (ASC) \cite{tsakiris2017ASC}, K-subspaces (KSS) \cite{bradley2000k,agarwal2004}, etc. Here we focus on a robust HC model called DPCP-KSS \cite{ding2021dual} combined with the KSS method. Dual Principal Component Pursuit (DPCP) \cite{tsakiris2015} is a non-convex method that is specifically designed for robustly learning a single high-dimensional subspace, such as a hyperplane.

For learning a single hyperplane, the DPCP \cite{tsakiris2015} method aims to robustly find a normal vector to a hyperplane by solving:
\begin{align}\label{DPCP}
  \min_{\mathbf{b}\in \mathbb{S}^{D-1}}f(\mathbf{v}):=\|\tilde{X}^T\mathbf{v}\|_1, s.t.  \|\mathbf{v}\|_2=1,
\end{align}
where the dataset $\tilde{X}\in \mathbb{R}^{D\times (N+M)}$ is $\ell_2$ column-normalized, $\mathbb{S}$ is a $(D-1)$-dimensional unit sphere, $M$ denotes the number of outliers. Ding et al. \cite{ding2021dual} combined the KSS method, extended DPCP to incorporate multiple hyperplane clustering, and applied it to the plane detection task in point cloud data. They iteratively recovered $K$ normal vectors of hyperplanes using the Riemannian subgradient method (RSGM) and $K$ initial normal vectors $\mathbf{v}_k^0$. When applying DPCP-KSS, it is necessary to convert the sample points into homogeneous coordinates by adding 1 in advance, and then normalize them to the unit sphere; more details can be seen in \cite{ding2021dual}.

\section{The proposed method} \label{sec3}
\subsection{RFLkPC model}
This section introduces our RFLkPC model, a novel robust approach to improve the performance of fuzzy plane clustering. Inspired by the work of the LkF model by Wang et al. \cite{wang2011}, we generalize the local idea to fuzzy plane clustering. In addition, in order to better alleviate the effect of outliers, we use a mixture of hinge loss and $L_1$ norm as our distance function to improve the robustness of FkPC \cite{zhu2008}. The objective function of the proposed RFLkPC  model is as follows:
\begin{align}\label{RFLkPC}
  \min_{u_{ik},\mathbf{v_k},\mathbf{\mu_k}}  &\sum_{k=1}^{K}\sum_{i=1}^{N} u_{ik}^m \left(\alpha\|\mathbf{v_k}^T(x_i-\mathbf{\mu_k})\|_2^2 \quad + \right. \\
  \nonumber &\left.(1-\alpha)\|\mathbf{v_k}^T(x_i-\mathbf{\mu_k})\|_1+
    \lambda\|x_i-\mathbf{\mu_k}\|_2^2 \right)\\
  \nonumber & s.t.  \|\mathbf{v_k}\|_2 =1,\sum_{k=1}^{K}u_{ik}=1,u_{ik}\in [0,1],
\end{align}
where m > 1 is the fuzzification factor to control the fuzziness of the partition, $\mathbf{v_k}\in \mathbb{R}^{D\times 1}$ denotes the normal vector of the k-th plane cluster, $\mathbf{\mu_k}\in \mathbb{R}^{D\times 1}$ is the center of each plane cluster. The weight coefficient $\alpha\in [0,1]$ is used to balance the mixing distance, and $\lambda>0$ controls the localized degree to the center of the plane.

\begin{figure}[!t] 
  \centering
  \includegraphics[width=0.83\columnwidth]{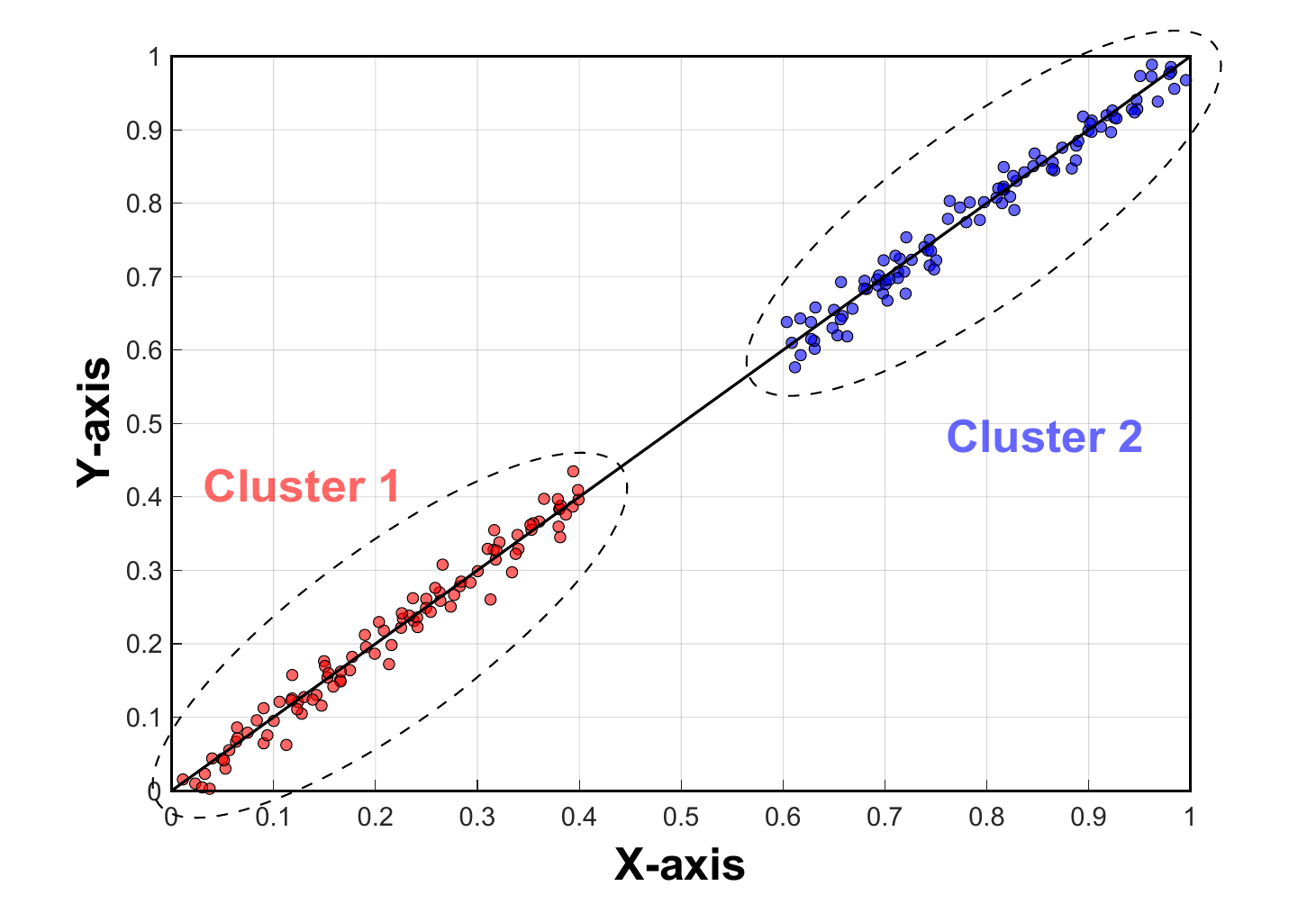}
  \caption{Line clusters may be too expanding!}
  \label{fig-lineE}
\end{figure}

RFLkPC contains two core ideas, one is a locally bounded constraint on plane clusters, and the other is a mixture of distance that characterizes projections to plane clusters. On the one hand, many early works \cite{bezdek1981,wang2011,yang2015local} realized that the assumption of infinitely expanding plane clusters could be detrimental to plane clustering. For example, Bezdek \cite{bezdek1981} pointed out that since lines have infinite length, it may be the case that collinear clusters which are in fact widely separated will not be identified as shown in Fig. \ref{fig-lineE}.

\begin{figure}[!t] 
  \centering
  \includegraphics[width=0.86\columnwidth]{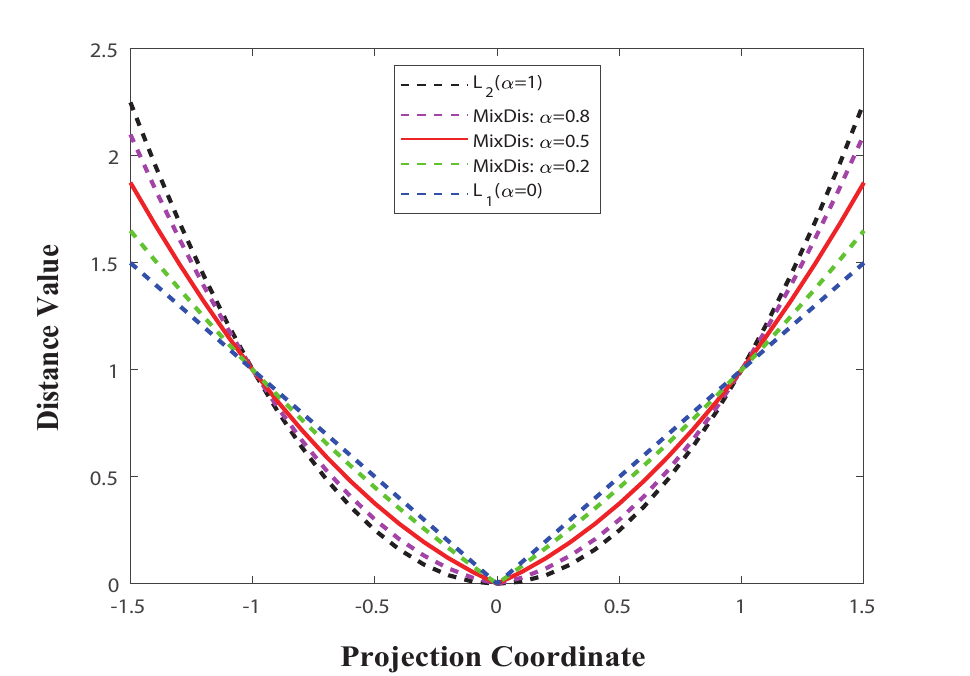}
  \caption{Illustration of mixture distance, $L_1$ and $L_2$ distance.}
  \label{fig-mixdis}
\end{figure}

On the other hand, ignoring the bounded term temporarily, from Fig. \ref{fig-mixdis} we can observe the connection between the mixture distance (projection of samples to plane clusters), and the common $L_2$ or $L_1$ distance. The mixture distance is a generalized distance, and $L_2$ or $L_1$ is a special case of it ($\alpha=1$ or 0). When the parameter $\alpha$ in the mixture distance is smaller, the growth rate of the distance value is first higher and then lower. This characteristic enables it to reduce the impact of noise or outliers. Conversely, when $\alpha$ is larger, the characteristic is opposite. Therefore, the RFLkPC model can accommodate different plane clustering error cases by flexibly adjusting $\alpha$.

Also, note that RFLkPC has some associations with LkF \cite{wang2011}. When $\alpha=1$, RFLkPC degenerates into a fuzzy version of LkF ($d=D-1$). When $\alpha=0$, RFLkPC can be regarded as a generalized fuzzy robust $L_1$ model for localized $K$ plane clustering (LkF with $d=D-1$).

In practical applications, the choice of $\lambda$ and $\alpha$ values in RFLkPC can depend on the characteristics of the data and the desired clustering performance. When the clusters in the dataset are strongly bounded, it is recommended to use a larger value for $\lambda$. This helps to enforce stronger cluster boundaries and ensure that points within a cluster are more closely connected. On the other hand, when the cluster boundaries are larger or less well-defined, a smaller value for $\lambda$ can be used. This allows for more flexibility in the clustering process and allows points that are farther apart to still be considered part of the same cluster. When the data contains certain noise or outliers, it is recommended to set $\alpha$ to a value close to 0. This helps to reduce the influence of noise on the clustering process and improve the robustness of the algorithm. The default value for $\alpha$ in RFLkPC is typically set to 0.5, but it can be adjusted based on the specific characteristics of the dataset.

\subsection{The Optimization Algorithm of RFLkPC}
There are three variables to be updated in the RFLkPC model \eqref{RFLkPC}, including the membership degree matrix $U$, the plane normal vectors $\mathbf{v_k}$, and the plane centers $\mathbf{\mu_k}$. We will take an alternate optimization strategy to update each variable, respectively.
\subsubsection{Update U}
Define $D_{ik}=\alpha\|\mathbf{v_k}^T(x_i-\mathbf{\mu_k})\|_2^2+(1-\alpha)\|\mathbf{v_k}^T(x_i-
\mathbf{\mu_k})\|_1+\lambda\|x_i-\mathbf{\mu_k}\|_2^2$, when we fix $\mathbf{v_k}$ and $\mathbf{\mu_k}$ in Eq. \eqref{RFLkPC}, the update of $U$ is the same as the iterative formula of the classic FCM \cite{Bezdek1994}, as follows:
\begin{equation}\label{update-U}
  u_{ik}= \left(\frac{D_{ik}}{\sum_{c=1}^{K}D_{ic}}\right)^{\frac{-1}{m-1}}.
\end{equation}

\subsubsection{Update $\mathbf{v_k}$}
When we fix $U$ and $\mathbf{\mu_k}$, problem in Eq. \eqref{RFLkPC} is transformed into the following problem.
\begin{align}\label{RFLkPC2}
  \min_{\mathbf{v_k}}  &\sum_{k=1}^{K}\sum_{i=1}^{N} u_{ik}^m \left(\alpha\|\mathbf{v_k}^T(x_i-\mathbf{\mu_k})\|_2^2  + (1-\alpha)\|\mathbf{v_k}^T(x_i-\mathbf{\mu_k})\|_1\right)\\
  \nonumber & s.t.  \|\mathbf{v_k}\|_2 =1.
\end{align}

Eq. \eqref{RFLkPC2} can be further divided into $K$ similar subproblems and solve them separately. Let the auxiliary variable $s_{ik}=\frac{1}{\|\mathbf{v_k}^T(x_i-\mathbf{\mu_k})\|_1}$, the k-th subproblem is equivalent to the following problem:
\begin{align}\label{RFLkPC3}
  \min_{\mathbf{v_k}}  &\sum_{i=1}^{N} u_{ik}^m(\alpha+s_{ik}(1-\alpha))\|\mathbf{v_k}^T(x_i-\mathbf{\mu_k})\|_2^2\\
  \nonumber & s.t.  \|\mathbf{v_k}\|_2 =1.
\end{align}

The Lagrange function of Eq. \eqref{RFLkPC3} is as follows:
\begin{align}\label{RFLkPC4}
  \min_{\mathbf{v_k}} \sum_{i=1}^{N} u_{ik}^m(\alpha+s_{ik}(1-\alpha))\|\mathbf{v_k}^T(x_i-\mathbf{\mu_k})\|_2^2 -\xi_{k}(\|\mathbf{v_k}\|_2-1),
\end{align}
where $\xi_{k}$ denotes the k-th Lagrange multiplier. When $s_{ik}$ is fixed, $g_{ik}=(\alpha+s_{ik}(1-\alpha))$, let the derivative of Eq. \eqref{RFLkPC4} with respect to $\mathbf{v_k}$ equal to zero, then we have:
\begin{align}\label{vk_deri}
 \sum_{i=1}^{N} u_{ik}^{m}g_{ik}(x_i-\mathbf{\mu_k})(x_i-\mathbf{\mu_k})^T\mathbf{v_k} -\xi_{k}\mathbf{v_k}=0.
\end{align}

Let $W_{k}=\sum_{i=1}^{N} u_{ik}^{m}g_{ik}(x_i-\mathbf{\mu_k})(x_i-\mathbf{\mu_k})^T$, then:
\begin{align}\label{vk_update}
 W_{k}\mathbf{v_k} =\xi_{k}\mathbf{v_k}.
\end{align}

Here $W_{k} \in \mathbb{R}^{D\times D}$, referring to \cite{bradley2000k}, we can easily derive that $\xi_{k}$ is the smallest eigenvalue of $W_{k}$, and $\mathbf{v_k}$ is the eigenvector of $W_{k}$ corresponding to the smallest eigenvalue of $W_{k}$. The solution $\mathbf{v_k}$ of \eqref{vk_update} can be obtained through singular value decomposition (SVD).

\subsubsection{Update $\mathbf{\mu_k}$}
When we fix $U$ and $\mathbf{v_k}$, the optimization of $\mathbf{\mu_k}$ in Eq. \eqref{RFLkPC} is similar to that of $\mathbf{v_k}$. Still based on the auxiliary variable $s_{ik}$ and $g_{ik}$ from the previous section, the optimization of the k-th plane center $\mathbf{\mu_k}$ is as follows:
\begin{align}\label{RFLkPC-muk}
 \min_{\mathbf{\mu_k}} \sum_{i=1}^{N} u_{ik}^{m}(g_{ik}\|\mathbf{v_k}^T(x_i-\mathbf{\mu_k})\|_2^2 + \lambda\|x_i-\mathbf{\mu_k}\|_2^2).
\end{align}

Let the derivative of the objective function in \eqref{RFLkPC-muk} with respect to $\mathbf{\mu_k}$ equal to zero, then we have:
\begin{align}\label{muk_deri}
 \sum_{i=1}^{N} u_{ik}^{m}(g_{ik}\mathbf{v_k}\mathbf{v_k}^T(x_i-\mathbf{\mu_k}) +\lambda(x_i-\mathbf{\mu_k}))=0.
\end{align}
\begin{align}\label{muk_deri2}
 \sum_{i=1}^{N} u_{ik}^{m}(g_{ik}\mathbf{v_k}\mathbf{v_k}^T+\lambda I)\mathbf{\mu_k}=\sum_{i=1}^{N} u_{ik}^{m}(g_{ik}\mathbf{v_k}\mathbf{v_k}^T+\lambda I)x_i.
\end{align}

Define $B_k=\sum_{i=1}^{N} u_{ik}^{m}(g_{ik}\mathbf{v_k}\mathbf{v_k}^T+\lambda I)$, where $I$ is $D$-order unit matrix. Since $g_{ik}>0$ and $\lambda>0$, it is easy to know that $B_k$ is an invertible matrix. Therefore, the solution of $\mathbf{\mu_k}$ in \eqref{muk_deri2} is to solve a system of linear equations of a $D$-order matrix ($B_k\in \mathbb{R}^{D\times D}$) with Gaussian elimination, and the format is as follows:
\begin{align}\label{muk_update}
 \mathbf{\mu_k}=B_{k}^{-1}\left(\sum_{i=1}^{N} u_{ik}^{m}(g_{ik}\mathbf{v_k}\mathbf{v_k}^T+\lambda I)x_i\right).
\end{align}

Since the update of $\mathbf{v_k}$, $\mathbf{\mu_k}$ requires fixing the auxiliary variables $s_{ik}$ and $g_{ik}$, their optimization is based on an iterative re-weighting algorithm. The overall optimization process of the RFLkPC model \eqref{RFLkPC} is presented in Algorithm \ref{RFLkPC-Algorithm}.
\begin{algorithm}[H]
\caption{Iterative re-weighting optimization for RFLkPC.}
\label{RFLkPC-Algorithm}
\begin{algorithmic}[1]
\STATE \textbf{Input:} the datasets $\{x_1,\dots,x_N\}\subset R^{D}$, the number of clusters $K$, the fuzzy factor $m$, the parameter $\lambda$ and $\alpha$, the termination error $\eta$.
\STATE \textbf{Initialization:} plane normal vectors $\mathbf{v_k}$, plane centers $\mathbf{\mu_k}$.	
\STATE \textbf{while} not converge \textbf{do}
\STATE \quad Update $U$ by \eqref{update-U}.
\STATE \quad \textbf{while} not converge \textbf{do}
\STATE \quad \quad Update $s_{ik}$, $g_{ik}$ with $U$ and $\mathbf{v_k}$.
\STATE \quad \quad Update $\mathbf{v_k}$ by \eqref{vk_update}.
\STATE \quad \quad Update $\mathbf{\mu_k}$ by \eqref{muk_update}.
\STATE \quad \textbf{end while}.
\STATE \textbf{end while}.
\STATE \textbf{Output:} the membership $U_{ik}$, plane normal vectors $\mathbf{v_k}$ and plane centers $\mathbf{\mu_k}$.
\end{algorithmic}
\end{algorithm}
The converge criterion of the outer loop is to observe whether the iterative difference (Frobenius-norm measure) of $U$ is less than $\eta$. The termination of the inner loop can be determined by the iterative difference (Frobenius-norm measure) of plane normal vectors $V(\mathbf{v_k})$ or the maximum number of iterations. The computational complexity of updating $U$ in \eqref{update-U} is $\mathcal{O}(NKD)$. The computational complexity of updating $s_{ik}$ and $g_{ik}$ is $\mathcal{O}(NKD)$. The computational complexity of calculating $W_k$ in \eqref{vk_update} is $\mathcal{O}(ND^3)$. Therefore,the computational complexity of updating $K$ normal vectors $\mathbf{v_k}$ in \eqref{vk_update} is $\mathcal{O}(NKD^3+KD^3)$ by eigenvalue decomposition. The computational complexity of calculating $B_k$ is $\mathcal{O}(ND^3)$. The computational complexity of updating $\mathbf{\mu_k}$ in \eqref{muk_update} is totally $\mathcal{O}(NKD^3+KD^2)$ by Gaussian elimination. When the number of converging steps of the inner loop is $T$, the total computational complexity of Algorithm \ref{RFLkPC-Algorithm} in each iteration is $\mathcal{O}((T+1)NKD+2TNKD^3+TKD^3+TKD^2)$, which approximates $\mathcal{O}(TNKD^3)$ ignoring the low order term.

\begin{figure*}[!h]
	\centering
	\begin{minipage}{0.3\linewidth}
		\centering
		\includegraphics[width=0.98\linewidth]{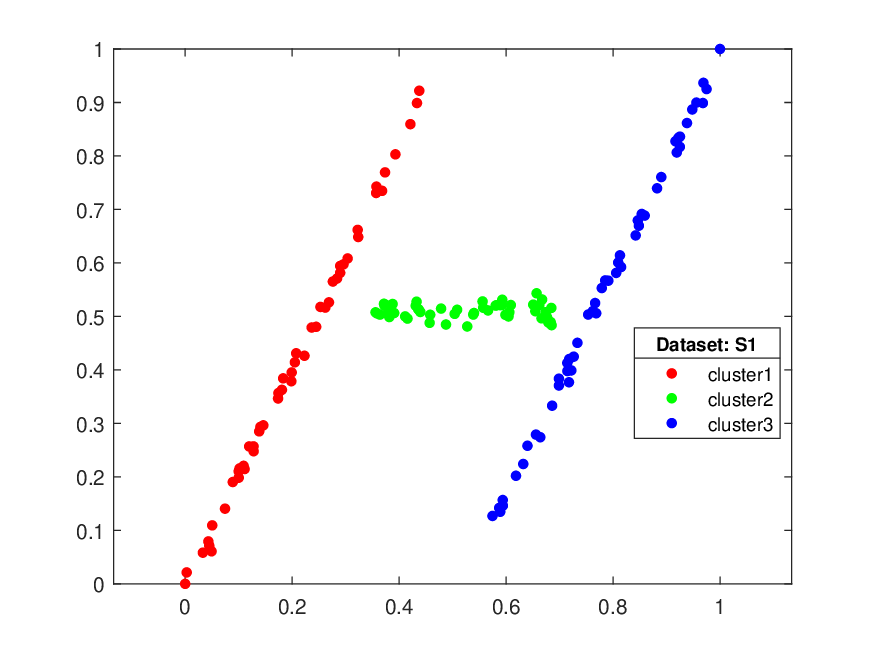}
	\end{minipage}
	\begin{minipage}{0.3\linewidth}
		\centering
		\includegraphics[width=0.98\linewidth]{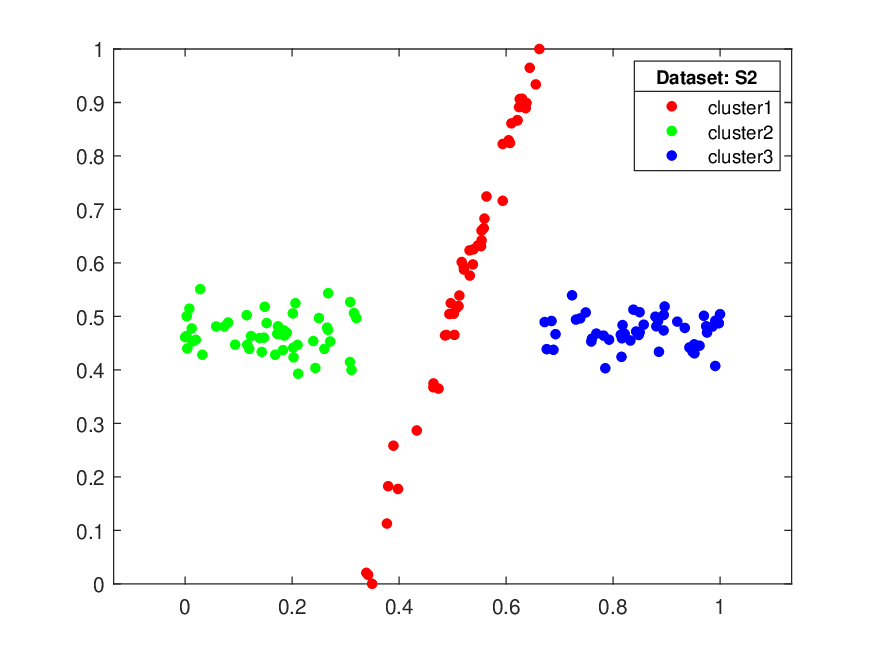}
	\end{minipage}
	\begin{minipage}{0.3\linewidth}
		\centering
		\includegraphics[width=0.98\linewidth]{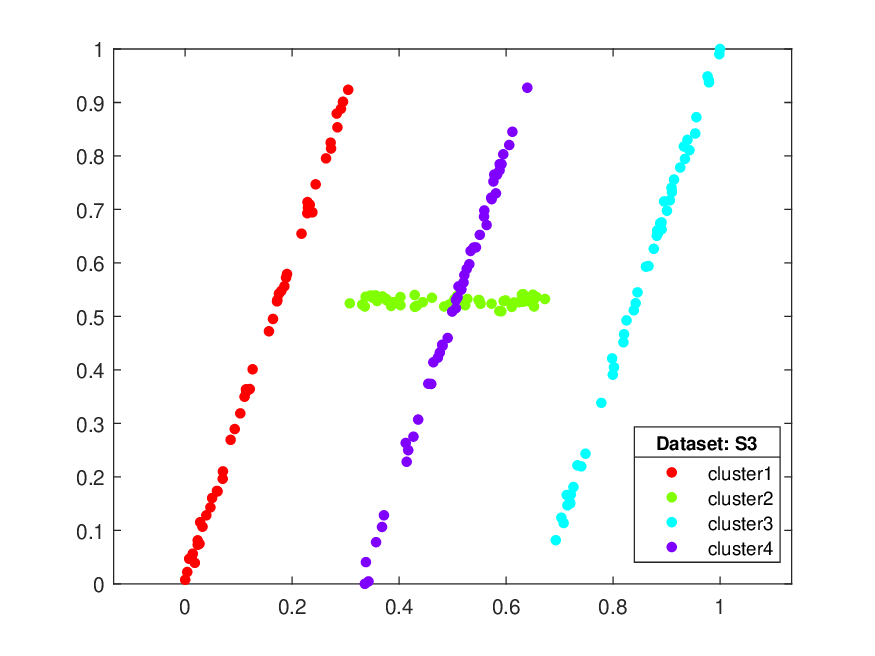}
	\end{minipage}
	
	\begin{minipage}{0.3\linewidth}
		\centering
		\includegraphics[width=0.98\linewidth]{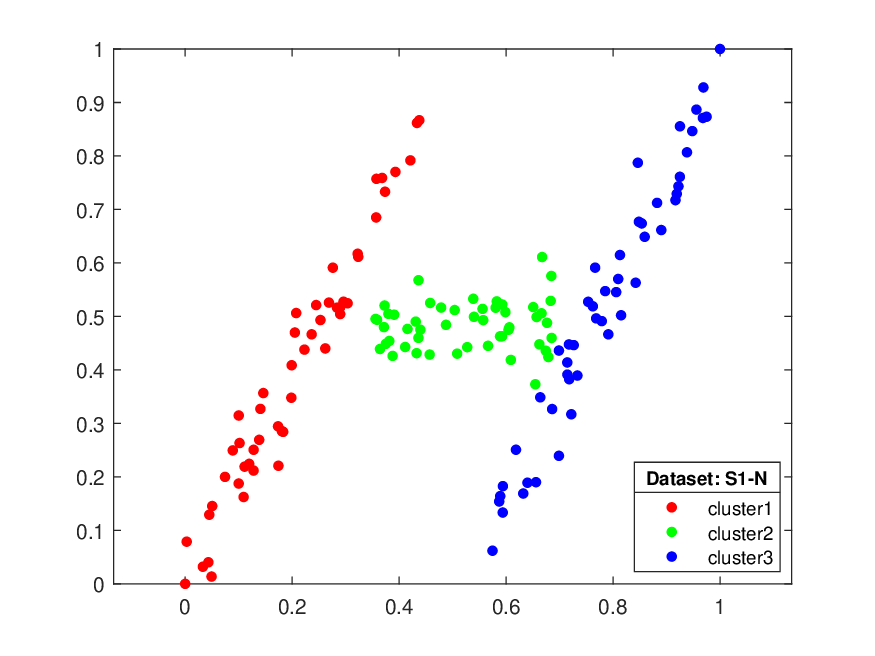}
	\end{minipage}
	\begin{minipage}{0.3\linewidth}
		\centering
		\includegraphics[width=0.98\linewidth]{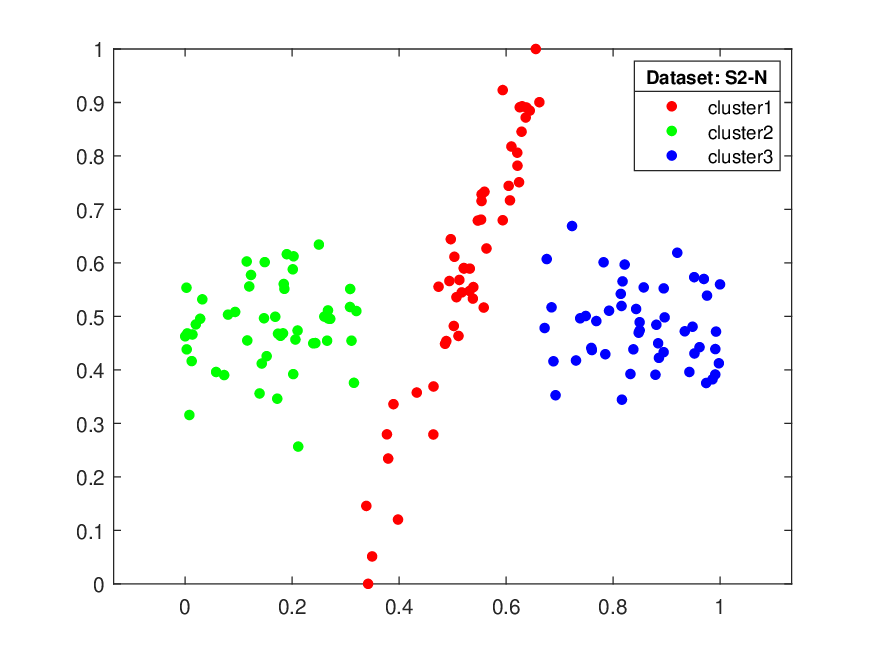}
	\end{minipage}
	\begin{minipage}{0.3\linewidth}
		\centering
		\includegraphics[width=0.98\linewidth]{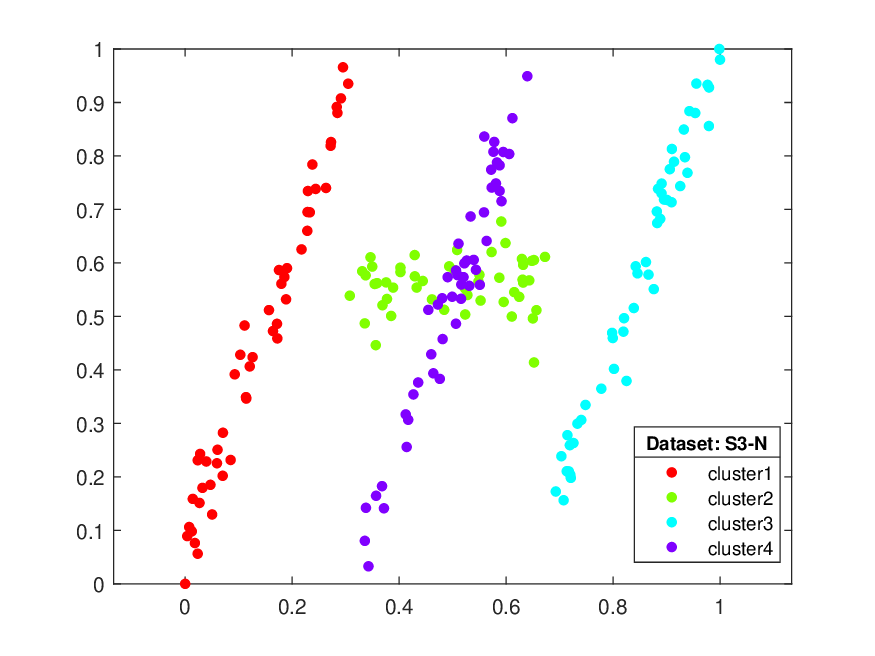}
	\end{minipage}
	
	\begin{minipage}{0.3\linewidth}
		\centering
		\includegraphics[width=0.98\linewidth]{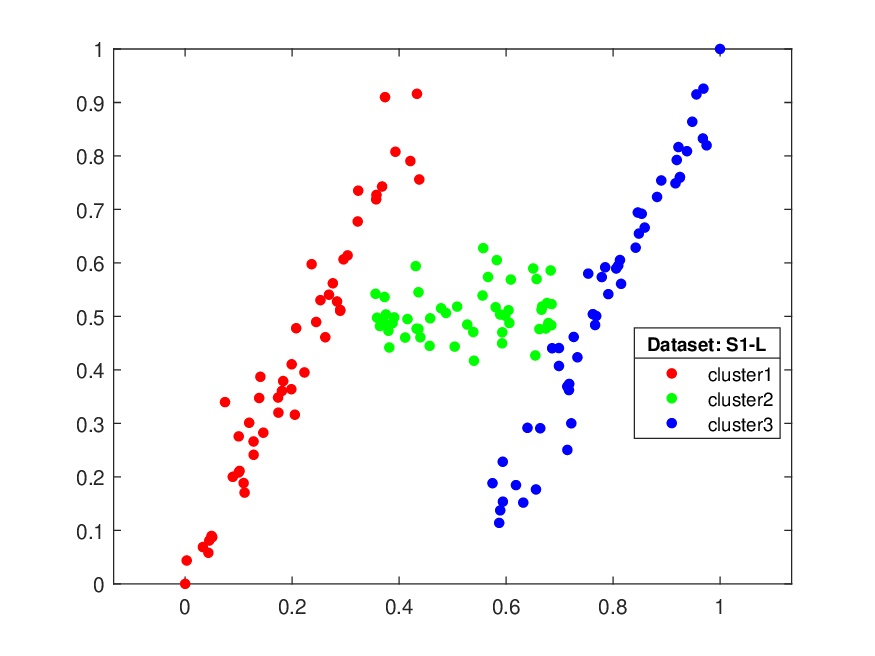}
	\end{minipage}
	\begin{minipage}{0.3\linewidth}
		\centering
		\includegraphics[width=0.98\linewidth]{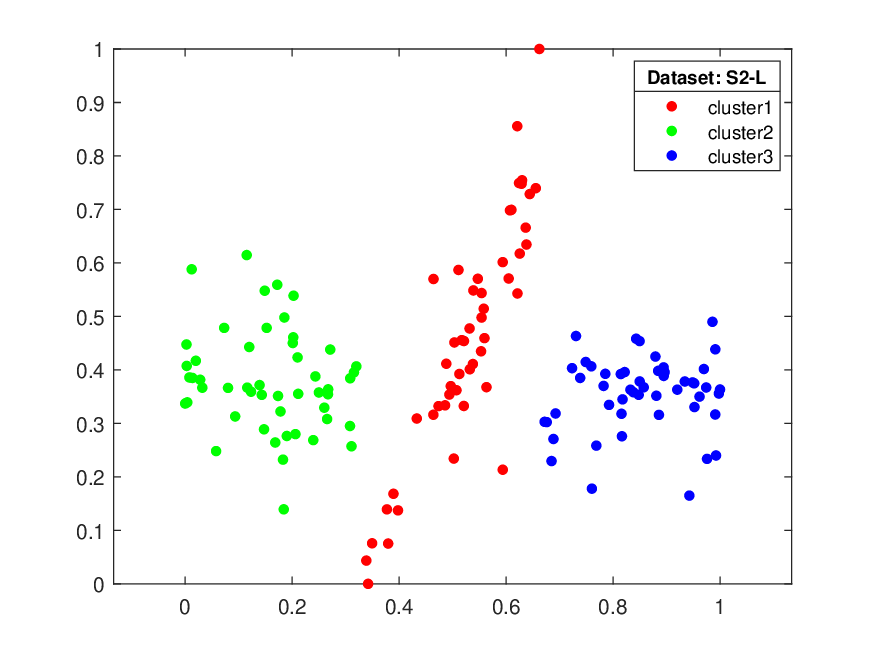}
	\end{minipage}
	\begin{minipage}{0.3\linewidth}
		\centering
		\includegraphics[width=0.98\linewidth]{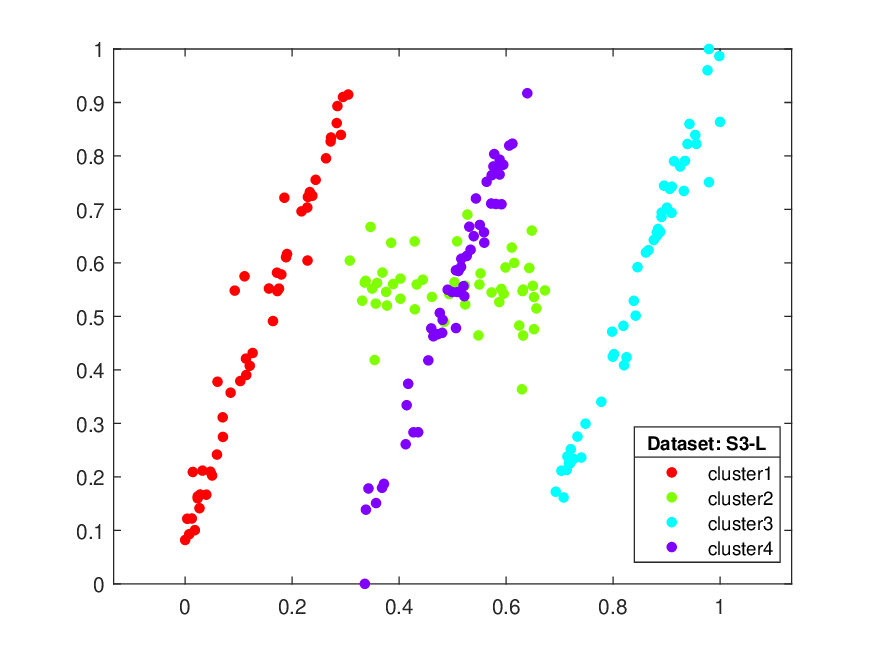}
	\end{minipage}
	
	\begin{minipage}{0.3\linewidth}
		\centering
		\includegraphics[width=0.98\linewidth]{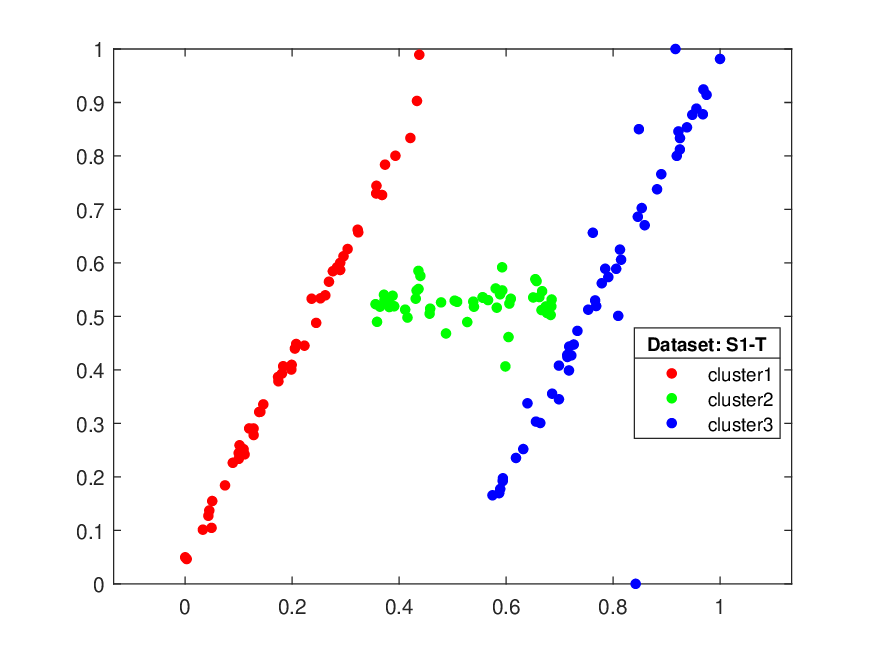}
	\end{minipage}
	\begin{minipage}{0.3\linewidth}
		\centering
		\includegraphics[width=0.98\linewidth]{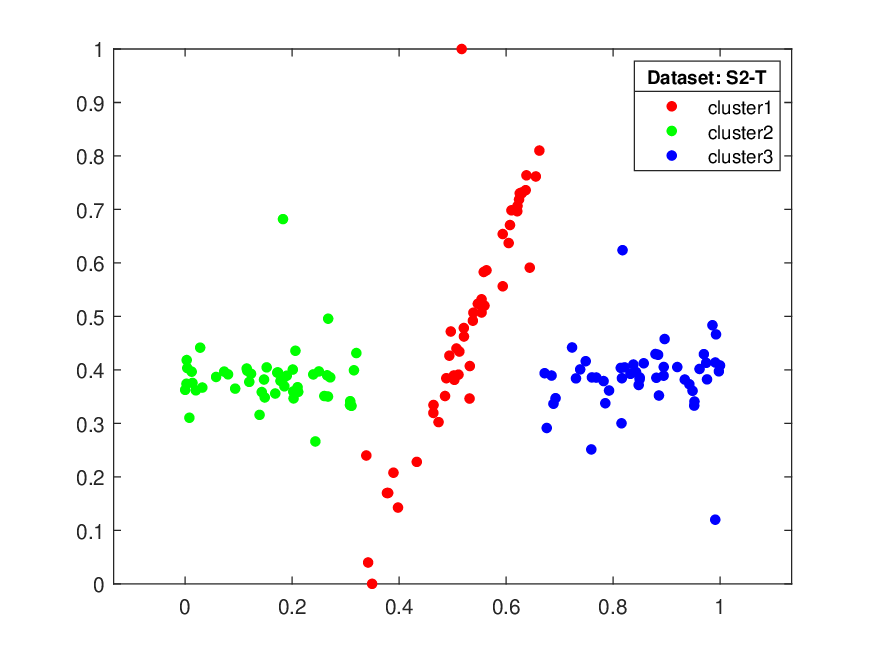}
	\end{minipage}
	\begin{minipage}{0.3\linewidth}
		\centering
		\includegraphics[width=0.98\linewidth]{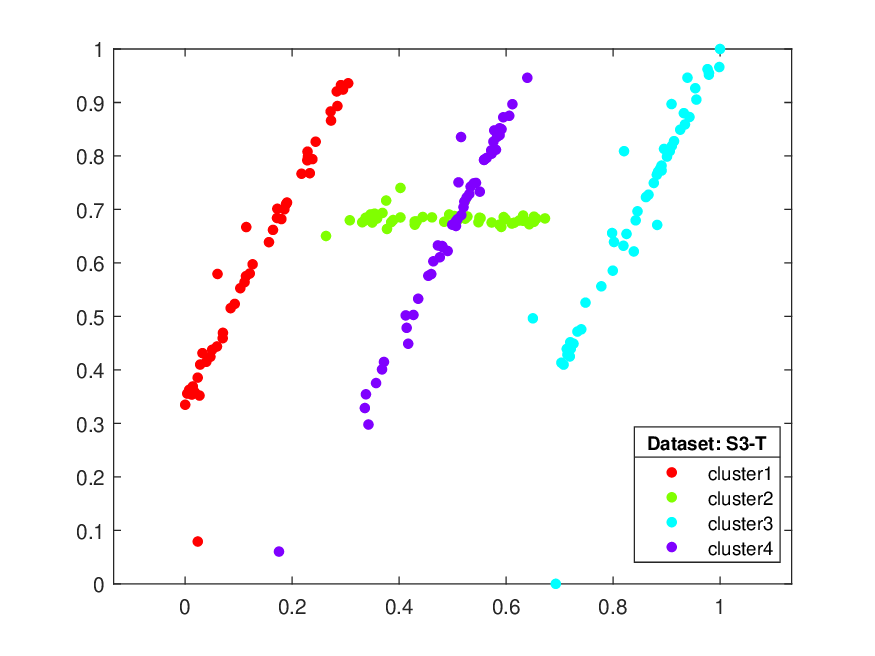}
	\end{minipage}

	\begin{minipage}{0.3\linewidth}
		\centering
		\includegraphics[width=0.98\linewidth]{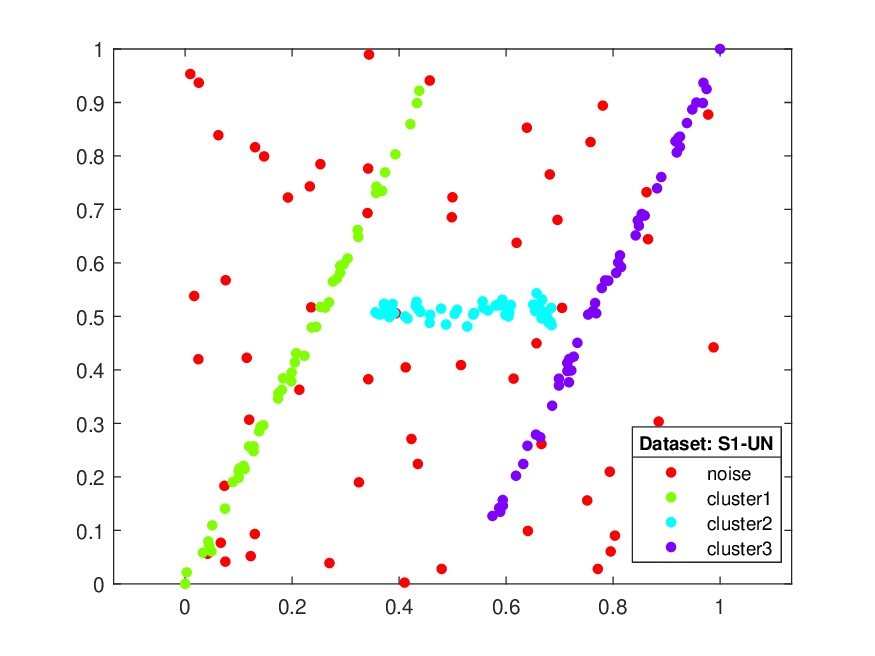}
	\end{minipage}
	\begin{minipage}{0.3\linewidth}
		\centering
		\includegraphics[width=0.98\linewidth]{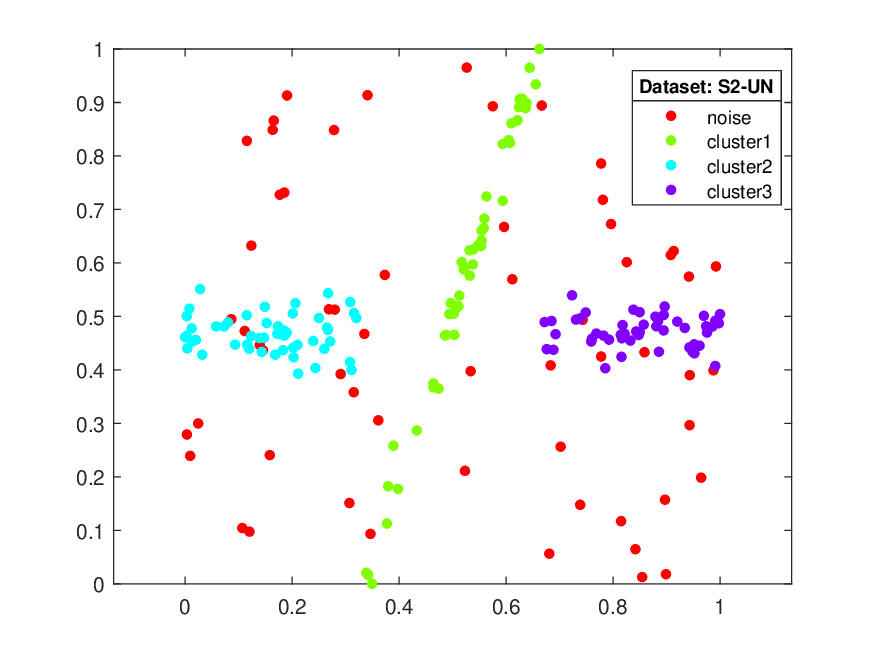}
	\end{minipage}
	\begin{minipage}{0.3\linewidth}
		\centering
		\includegraphics[width=0.98\linewidth]{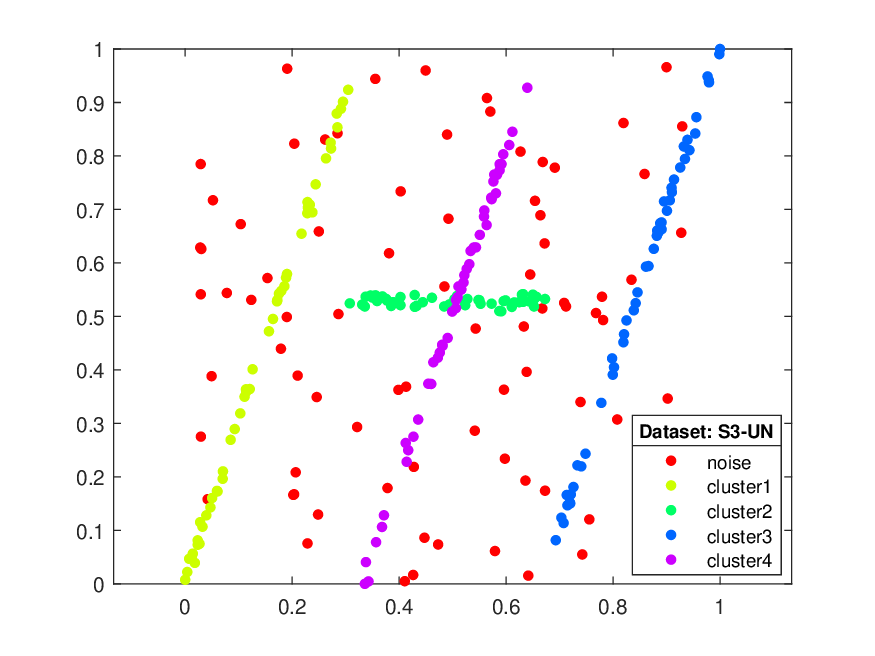}
	\end{minipage}
    \caption{The simulated datasets:three types of plane clustering with different noise or outliers. The dataset Si ($i=1,2,3$) contains no outliers, while the noise in Si-N follows the normal distribution, Si-L follows the Laplace distribution, and Si-T follows the t-distribution. Si-UN is obtained by uniformly adding 40\% noises or outliers into Si (noises in Si-UN marked by red dots).}
    \label{fig-Si}
\end{figure*}

\begin{table*} [h!]
\centering
\caption{{Mean scores of Various Methods on synthetic Datasets.}}
\label{Tab-Si-score}
    \resizebox{1.95\columnwidth}{!}{
\begin{tabular*}{1.08\linewidth}{>{}l >{}l >{}l >{}l >{}l >{}l >{}l >{}l >{}l >{}l >{}l}
    \hline
    Dataset & KPC & FkPC & LkPPC & SCC & DPCP-KSS & FCRM &FWCHR &LFDC &SNMoE &RFLkPC\\
    \hline
    \multirow{4}*{~} &ACC(\%) &ACC(\%) &ACC(\%) &ACC(\%) &ACC(\%) &ACC(\%) &ACC(\%) &ACC(\%) &ACC(\%) &ACC(\%)\\
    ~ &NMI(\%) &NMI(\%) &NMI(\%) &NMI(\%) &NMI(\%) &NMI(\%) &NMI(\%) &NMI(\%) &NMI(\%) &NMI(\%)\\
~ &ARI(\%) &ARI(\%) &ARI(\%) &ARI(\%) &ARI(\%) &ARI(\%) &ARI(\%) &ARI(\%) &ARI(\%) &ARI(\%)\\
~ &Purity(\%) &Purity(\%) &Purity(\%) &Purity(\%) &Purity(\%) &Purity(\%) &Purity(\%) &Purity (\%) &Purity(\%) &Purity(\%)\\
    \hline

 & 95.57$\pm$0.63 & 98.00$\pm$0.00 & 96.77$\pm$0.75 & 70.65$\pm$0.19 & 97.33$\pm$0.00 & 96.00$\pm$0.00 & 96.00$\pm$0.00 & \underline{98.57$\pm$0.05} & 95.75$\pm$0.34 & \textbf{\textbf{100.00$\pm$0.00}} \\
S1 & 88.15$\pm$0.98 & 91.88$\pm$0.00 & \underline{94.18$\pm$1.35} & 47.96$\pm$0.34 & 90.05$\pm$0.00 & 86.26$\pm$0.00 & 86.26$\pm$0.00 & 93.77$\pm$0.17 & 89.81$\pm$0.83 & \textbf{100.00$\pm$0.00} \\
N(150) & 89.17$\pm$1.26 & 93.98$\pm$0.00 & 93.22$\pm$1.57 & 33.11$\pm$0.47 & 92.04$\pm$0.00 & 88.10$\pm$0.00 & 88.10$\pm$0.00 & \underline{95.67$\pm$0.16} & 88.42$\pm$0.94 & \textbf{100.00$\pm$0.00} \\
& 95.57$\pm$0.63 & 98.00$\pm$0.00 & 96.77$\pm$0.75 & 70.65$\pm$0.19 & 97.33$\pm$0.00 & 96.00$\pm$0.00 & 96.00$\pm$0.00 & \underline{98.57$\pm$0.05} & 95.75$\pm$0.34 & \textbf{100.00$\pm$0.00} \\

\hline
&79.34$\pm$0.28 & 71.71$\pm$0.20 & \textbf{100.00$\pm$0.00} & 89.72$\pm$0.63 & 74.97$\pm$0.06 & 73.10$\pm$0.05 & 75.74$\pm$0.20 & \underline{98.14$\pm$0.46} & 95.01$\pm$0.35 & \textbf{100.00$\pm$0.00} \\
S2  & 68.36$\pm$0.25 & 56.02$\pm$0.03 & \textbf{100.00$\pm$0.00} & 80.49$\pm$0.56 & 64.58$\pm$0.07 & 52.56$\pm$0.26 & 57.97$\pm$0.11 & \underline{94.42$\pm$0.66} & 88.54$\pm$0.79 & \textbf{100.00$\pm$0.00} \\
N(150)  & 57.64$\pm$0.26 & 49.03$\pm$0.07 & \textbf{100.00$\pm$0.00} & 75.94$\pm$0.92 & 53.50$\pm$0.05 & 46.91$\pm$0.11 & 49.02$\pm$0.14 & \underline{95.41$\pm$0.82} & 86.62$\pm$0.93 & \textbf{100.00$\pm$0.00} \\
& 79.34$\pm$0.28 & 71.71$\pm$0.20 & \textbf{100.00$\pm$0.00} & 89.72$\pm$0.63 & 74.97$\pm$0.06 & 73.10$\pm$0.05 & 75.74$\pm$0.20 & \underline{98.14$\pm$0.46} & 95.01$\pm$0.35 & \textbf{100.00$\pm$0.00} \\

\hline
 & 74.74$\pm$1.03 & 79.21$\pm$2.04 & 75.93$\pm$0.96 & 55.48$\pm$0.38 & 75.94$\pm$1.31 & 51.50$\pm$1.49 & 55.08$\pm$1.06 & \textbf{85.60$\pm$0.98} & 78.28$\pm$1.83 & \underline{84.10$\pm$1.56} \\
S3 & 65.82$\pm$1.18 & 71.53$\pm$2.61 & 72.75$\pm$1.31 & 47.42$\pm$0.57 & 67.89$\pm$1.69 & 29.99$\pm$2.20 & 33.89$\pm$1.85 & \textbf{81.82$\pm$1.06} & 74.65$\pm$1.78 & \underline{77.25$\pm$1.89} \\
N(200) & 58.08$\pm$1.50 & 67.30$\pm$3.10 & 63.30$\pm$1.59 & 29.48$\pm$0.50 & 60.53$\pm$1.93 & 23.79$\pm$2.17 & 28.44$\pm$1.57 & \textbf{76.52$\pm$1.50} & 65.72$\pm$2.68 & \underline{74.10$\pm$2.36} \\
& 75.18$\pm$0.97 & 79.34$\pm$2.03 & 75.93$\pm$0.96 & 59.92$\pm$0.42 & 76.36$\pm$1.25 & 52.25$\pm$1.45 & 55.86$\pm$1.10 & \textbf{85.63$\pm$0.98} & 80.28$\pm$1.58 & \underline{84.70$\pm$1.46} \\

\hline
& 94.07$\pm$0.23 & \underline{94.43$\pm$0.03} & 93.31$\pm$0.20 & 70.89$\pm$0.32 & 92.64$\pm$0.07 & 89.01$\pm$0.06 & 80.64$\pm$2.02 & 84.41$\pm$1.04 & 86.98$\pm$0.41 & \textbf{96.52$\pm$0.03} \\
 S1-N  & 79.81$\pm$0.46 & \underline{80.53$\pm$0.08} & 78.76$\pm$0.34 & 49.70$\pm$0.40 & 76.64$\pm$0.10 & 69.79$\pm$0.11 & 58.91$\pm$2.74 & 65.23$\pm$1.32 & 73.18$\pm$0.80 & \textbf{86.68$\pm$0.09} \\
N(150) & 83.16$\pm$0.50 & \underline{83.95$\pm$0.09} & 80.82$\pm$0.44 & 37.21$\pm$0.45 & 78.77$\pm$0.20 & 68.73$\pm$0.16 & 58.46$\pm$2.78 & 62.82$\pm$1.86 & 66.96$\pm$0.98 & \textbf{89.84$\pm$0.08} \\
& 94.07$\pm$0.23 & \underline{94.43$\pm$0.03} & 93.31$\pm$0.20 & 70.89$\pm$0.32 & 92.64$\pm$0.07 & 89.01$\pm$0.06 & 80.64$\pm$2.02 & 84.41$\pm$1.04 & 86.98$\pm$0.41 & \textbf{96.52$\pm$0.03} \\

\hline
 & 76.00$\pm$0.00 & 74.63$\pm$0.24 & \underline{98.67$\pm$0.00} & 93.72$\pm$0.30 & 75.27$\pm$0.45 & 62.58$\pm$0.46 & 70.13$\pm$0.38 & 96.36$\pm$0.11 & 92.44$\pm$0.19 & \textbf{99.33$\pm$0.00} \\
 S2-N &58.99$\pm$0.31 & 56.97$\pm$0.34 & \underline{94.88$\pm$0.00} & 83.25$\pm$0.74 & 60.25$\pm$0.52 & 38.11$\pm$0.30 & 48.87$\pm$0.67 & 87.47$\pm$0.28 & 83.06$\pm$0.28 & \textbf{97.02$\pm$0.00} \\
N(150) & 51.14$\pm$0.07 & 48.82$\pm$0.18 & \underline{96.03$\pm$0.00} & 82.90$\pm$0.81 & 51.48$\pm$0.45 & 30.07$\pm$0.35 & 39.05$\pm$0.53 & 89.46$\pm$0.29 & 79.99$\pm$0.42 & \textbf{97.99$\pm$0.00} \\
& 76.00$\pm$0.00 & 74.63$\pm$0.24 & \underline{98.67$\pm$0.00} & 93.72$\pm$0.30 & 75.27$\pm$0.45 & 62.58$\pm$0.46 & 70.13$\pm$0.38 & 96.36$\pm$0.11 & 92.44$\pm$0.19 & \textbf{99.33$\pm$0.00} \\
\hline

& 77.70$\pm$1.39 & \underline{84.77$\pm$1.42} & 79.12$\pm$0.94 & 55.81$\pm$0.38 & 77.24$\pm$1.32 & 41.58$\pm$0.56 & 40.77$\pm$0.43 & 79.63$\pm$0.62 & 61.30$\pm$1.51 & \textbf{88.10$\pm$1.22} \\
S3-N  & 66.89$\pm$1.52 & 73.36$\pm$1.80 & \underline{75.20$\pm$0.96} & 48.21$\pm$0.64 & 65.26$\pm$1.59 & 14.72$\pm$0.72 & 15.10$\pm$0.67 & 72.13$\pm$0.72 & 60.76$\pm$1.33 & \textbf{76.40$\pm$1.58} \\
N(200) & 62.39$\pm$1.90 & \underline{71.12$\pm$1.97} & 67.28$\pm$1.27 & 31.80$\pm$0.72 & 60.25$\pm$1.93 & 9.57$\pm$0.56  & 10.31$\pm$0.52 & 65.24$\pm$0.85 & 43.13$\pm$1.94 & \textbf{75.14$\pm$1.67}\\
& 77.86$\pm$1.38 & \underline{84.80$\pm$1.41} & 79.22$\pm$0.92 & 61.15$\pm$0.43 & 77.63$\pm$1.29 & 42.28$\pm$0.53 & 42.81$\pm$0.46 & 79.75$\pm$0.59 & 66.75$\pm$1.11 & \textbf{88.23$\pm$1.17} \\

\hline

 & 89.98$\pm$0.76 & 90.79$\pm$0.73 & 91.28$\pm$0.92 & 73.50$\pm$0.31 & \textbf{96.60$\pm$0.03} & 92.07$\pm$0.02 & 89.52$\pm$0.96 & 94.65$\pm$0.13 & 86.65$\pm$0.77 & \underline{95.67$\pm$0.42} \\
S1-L & 74.15$\pm$1.32 & 74.81$\pm$1.33 & 80.04$\pm$1.62 & 51.82$\pm$0.40 & \textbf{88.21$\pm$0.12} & 77.10$\pm$0.04 & 73.91$\pm$1.43 &83.08$\pm$0.35 & 71.40$\pm$1.49 & \underline{86.50$\pm$0.87} \\
N(150)& 75.38$\pm$1.53 & 76.78$\pm$1.51 & 79.10$\pm$2.00 & 38.96$\pm$0.44 & \textbf{90.09$\pm$0.09} & 76.77$\pm$0.06 & 73.22$\pm$1.64 & 84.30$\pm$0.37 & 65.05$\pm$1.97 & \underline{88.20$\pm$0.97} \\
& 89.98$\pm$0.76 & 90.79$\pm$0.73 & 91.28$\pm$0.92 & 73.55$\pm$0.29 & \textbf{96.60$\pm$0.03} & 92.07$\pm$0.02 & 89.52$\pm$0.96 & 94.65$\pm$0.13 & 86.65$\pm$0.77 & \underline{95.67$\pm$0.42} \\

\hline
 & 69.67$\pm$0.41 & 68.33$\pm$0.41 & \underline{99.33$\pm$0.00} & 95.70$\pm$0.25 & 78.50$\pm$0.15 & 61.83$\pm$0.55 & 67.00$\pm$1.34 & 94.13$\pm$0.59 & 99.13$\pm$0.14 & \textbf{100.00$\pm$0.00} \\
 S2-L & 47.85$\pm$0.01 & 44.15$\pm$0.08 & \underline{97.02$\pm$0.00} & 87.73$\pm$0.69 & 51.58$\pm$0.38 & 33.98$\pm$0.18 & 35.54$\pm$1.33 & 83.77$\pm$1.05 & 96.66$\pm$0.25 & \textbf{100.00$\pm$0.00} \\
N(150) & 42.71$\pm$0.13 & 40.36$\pm$0.12 & \underline{99.33$\pm$0.00} & 87.98$\pm$0.69 & 49.35$\pm$0.32 & 26.29$\pm$0.34 & 31.77$\pm$1.44 & 84.56$\pm$1.29 & 97.50$\pm$0.35 & \textbf{100.00$\pm$0.00} \\
& 69.67$\pm$0.41 & 68.33$\pm$0.41 & \underline{97.99$\pm$0.00} & 95.70$\pm$0.25 & 78.50$\pm$0.15 & 61.83$\pm$0.55 & 67.00$\pm$1.34 & 94.13$\pm$0.59 & 99.13$\pm$0.14 & \textbf{100.00$\pm$0.00} \\

\hline

 & \textbf{84.90$\pm$1.31} & 78.63$\pm$1.88 & 73.72$\pm$0.56 & 54.92$\pm$0.40 & 74.30$\pm$1.46 & 42.05$\pm$0.99 & 40.74$\pm$1.03 & 80.04$\pm$0.43 & 64.13$\pm$1.71 & \underline{84.18$\pm$1.63} \\
S3-L & \underline{72.50$\pm$1.69} & 63.34$\pm$2.36 & 70.32$\pm$0.95 & 47.18$\pm$0.68 & 59.13$\pm$1.88 & 15.57$\pm$1.28 & 14.83$\pm$1.28 & \textbf{74.83$\pm$0.49} & 63.75$\pm$1.55 & 72.13$\pm$2.06 \\
N(200) & \textbf{70.21$\pm$1.86} & 61.02$\pm$2.63 & 60.46$\pm$1.06 & 30.77$\pm$0.82 & 55.13$\pm$2.12 & 11.21$\pm$1.22 & 10.78$\pm$1.23 & 67.34$\pm$0.58 & 47.05$\pm$2.27 & \underline{69.82$\pm$2.30} \\
& \textbf{84.97$\pm$1.29} & 78.93$\pm$1.82 & 74.05$\pm$0.53 & 60.07$\pm$0.42 & 75.01$\pm$1.42 & 43.38$\pm$0.93 & 42.74$\pm$0.95 & 80.04$\pm$0.43 & 69.58$\pm$1.27 & \underline{84.22$\pm$1.61} \\

\hline

 & 82.85$\pm$1.50 & \underline{97.89$\pm$0.02} & 85.11$\pm$1.02 & 70.74$\pm$0.15 & 84.36$\pm$1.41 & 95.33$\pm$0.00 & 96.67$\pm$0.00 & 90.85$\pm$0.90 & 94.37$\pm$0.63 & \textbf{100.00$\pm$0.00} \\
S1-T & 65.17$\pm$2.35 & \underline{90.73$\pm$0.08} & 68.45$\pm$1.42 & 49.28$\pm$0.34 & 70.47$\pm$1.83 & 84.77$\pm$0.00 & 88.43$\pm$0.00 & 77.14$\pm$1.39 & 85.98$\pm$1.10 & \textbf{100.00$\pm$0.00} \\
N(150) & 63.56$\pm$2.43 & \underline{93.71$\pm$0.07} & 66.90$\pm$1.73 & 36.98$\pm$0.51 & 68.04$\pm$2.25 & 86.23$\pm$0.00 & 90.14$\pm$0.00 & 78.27$\pm$1.68 & 85.37$\pm$1.51 & \textbf{100.00$\pm$0.00} \\
& 83.33$\pm$1.41 & \underline{97.89$\pm$0.02} & 85.31$\pm$0.98 & 70.74$\pm$0.15 & 84.56$\pm$1.37 & 95.33$\pm$0.00 & 96.67$\pm$0.00 & 91.05$\pm$0.84 & 94.37$\pm$0.63 & \textbf{100.00$\pm$0.00} \\

\hline
 & 68.96$\pm$0.08 & 74.26$\pm$0.35 & \textbf{99.33$\pm$0.00} & 95.51$\pm$0.17 & 68.63$\pm$0.13 & 72.43$\pm$0.45 & 67.04$\pm$0.08 & \underline{98.75$\pm$0.12} & 95.77$\pm$0.17 & \textbf{99.33$\pm$0.00} \\
S2-T & 51.56$\pm$0.12 & 51.50$\pm$0.02 & \textbf{97.02$\pm$0.00} & 87.99$\pm$0.36 & 54.33$\pm$0.14 & 49.38$\pm$0.08 & 43.32$\pm$0.43 & \underline{95.41$\pm$0.37} & 88.38$\pm$0.40 & \textbf{97.02$\pm$0.00} \\
N(150) & 41.90$\pm$0.03 & 45.77$\pm$0.28 & \textbf{97.99$\pm$0.00} & 87.46$\pm$0.47 & 43.29$\pm$0.06 & 42.05$\pm$0.18 & 36.06$\pm$0.17 & \underline{96.32$\pm$0.35} & 88.13$\pm$0.48 & \textbf{97.99$\pm$0.00} \\
& 68.96$\pm$0.08 & 74.26$\pm$0.35 & \textbf{99.33$\pm$0.00} & 95.51$\pm$0.17 & 68.63$\pm$0.13 & 72.43$\pm$0.45 & 67.04$\pm$0.08 & \underline{98.75$\pm$0.12} & 95.77$\pm$0.17 & \textbf{99.33$\pm$0.00} \\

\hline

& 76.03$\pm$1.74 & \underline{82.03$\pm$2.33} & 72.80$\pm$1.01 & 62.65$\pm$1.02 & 77.25$\pm$2.18 & 59.05$\pm$2.50 & 59.95$\pm$2.03 & 79.15$\pm$0.58 & 73.03$\pm$2.19 & \textbf{89.90$\pm$1.74} \\
S3-T  & 68.68$\pm$1.90 & \underline{74.16$\pm$3.38} & 66.57$\pm$1.92 & 56.21$\pm$1.12 & 67.45$\pm$3.06 & 40.23$\pm$4.60 & 39.62$\pm$3.44 & 73.37$\pm$0.83 & 69.28$\pm$2.25 & \textbf{83.89$\pm$1.76} \\
N(200) & 61.36$\pm$2.53 & \underline{67.55$\pm$3.82} & 56.33$\pm$1.90 & 40.59$\pm$1.48 & 61.02$\pm$3.66 & 32.61$\pm$4.48 & 30.99$\pm$2.86 & 65.52$\pm$0.80 & 58.00$\pm$3.08 & \textbf{80.90$\pm$2.83} \\
& 76.55$\pm$1.64 & \underline{82.03$\pm$2.33} & 72.90$\pm$0.99 & 66.43$\pm$0.65 & 77.25$\pm$2.18 & 59.05$\pm$2.50 & 60.97$\pm$1.85 & 79.15$\pm$0.58 & 73.98$\pm$1.98 & \textbf{89.90$\pm$1.74} \\

\hline

 & 67.60$\pm$2.10 & 76.56$\pm$1.74 & 82.88$\pm$0.91 & 68.14$\pm$0.43 & 78.51$\pm$1.29 & 52.08$\pm$1.58 & 59.41$\pm$0.98 & \textbf{95.15$\pm$0.63} & 87.29$\pm$1.24 & \underline{91.65$\pm$0.14} \\
S1-UN & 45.96$\pm$3.17 & 54.20$\pm$2.86 & 67.23$\pm$1.41 & 48.92$\pm$0.46 & 58.22$\pm$1.99 & 20.25$\pm$2.52 & 26.69$\pm$1.74 & \textbf{87.73$\pm$0.95} & 79.94$\pm$1.59 & \underline{79.64$\pm$0.31} \\
 N(150+60)  & 23.54$\pm$1.64 & 28.98$\pm$1.40 & 32.89$\pm$1.00 & 20.81$\pm$0.26 & 30.21$\pm$1.09 & 8.93$\pm$1.33 & 12.88$\pm$0.85 & \textbf{47.90$\pm$0.63} & \underline{46.83$\pm$1.42} & 42.45$\pm$0.20 \\
 & 64.88$\pm$0.92 & 68.62$\pm$0.72 & 71.37$\pm$0.58 & 63.78$\pm$0.24 & 68.88$\pm$0.56 & 56.29$\pm$0.79 & 57.99$\pm$0.68 & \textbf{79.95$\pm$0.38} & 70.52$\pm$0.66 & \underline{77.11$\pm$0.13} \\

\hline

 & 88.63$\pm$1.38 & 84.00$\pm$1.85 & \textbf{100.00$\pm$0.00} & 87.23$\pm$0.77 & 87.87$\pm$1.67 & 45.33$\pm$0.00 & 63.24$\pm$0.70 & 87.46$\pm$1.62 & 96.09$\pm$0.62 & \underline{98.05$\pm$0.07} \\
 S2-UN & 81.47$\pm$1.69 & 68.00$\pm$2.78 & \textbf{100.00$\pm$0.00} & 79.09$\pm$0.52 & 80.57$\pm$2.21 & 29.16$\pm$0.00 & 35.12$\pm$0.58 & 79.01$\pm$2.27 & 91.44$\pm$0.94 & \underline{93.32$\pm$0.19} \\
N(150+60) & 43.31$\pm$1.12 & 37.36$\pm$1.68 & \textbf{53.82$\pm$0.00} & 39.28$\pm$0.47 & 43.32$\pm$1.39 & 8.21$\pm$0.00 & 18.86$\pm$0.36 & 42.46$\pm$1.44 & \underline{51.64$\pm$0.82} & 51.19$\pm$0.12 \\
& 71.33$\pm$0.85 & 69.95$\pm$0.82 & \textbf{80.95$\pm$0.00} & 70.00$\pm$0.36 & 70.38$\pm$0.97 & 52.38$\pm$0.00 & 64.59$\pm$0.27 & 69.72$\pm$0.90 & 71.55$\pm$0.41 & \underline{76.56$\pm$0.07} \\

\hline

& 63.43$\pm$1.10 & \underline{72.29$\pm$1.71} & 68.63$\pm$0.86 & 52.1$\pm$0.66 & 66.83$\pm$1.18 & 45.07$\pm$0.49 & 48.23$\pm$0.33 & 68.23$\pm$0.74 & 67.30$\pm$1.00 & \textbf{72.40$\pm$1.70} \\
S3-UN  &  51.68$\pm$1.94 & 61.37$\pm$2.31 & \underline{69.38$\pm$2.05} & 44.36$\pm$0.71 & 55.71$\pm$1.80 & 18.69$\pm$0.61 & 20.57$\pm$0.57 & 62.36$\pm$1.00 & 65.74$\pm$1.21 & \textbf{61.49$\pm$2.33} \\
N(200+80) & 23.65$\pm$1.01 & 29.39$\pm$1.43 & 28.56$\pm$1.09 & 15.02$\pm$0.42 & 25.70$\pm$0.98 & 7.55$\pm$0.30 & 9.22$\pm$0.29 & 26.99$\pm$0.64 & \textbf{31.58$\pm$0.91} & \underline{29.49$\pm$1.43} \\
& 57.35$\pm$0.65 & 60.81$\pm$0.99 & 58.59$\pm$0.57 & 53.82$\pm$0.28 & 58.26$\pm$0.63 & 49.52$\pm$0.21 & 50.68$\pm$0.36 & 60.29$\pm$0.29 & \textbf{63.11$\pm$0.43} & \underline{60.86$\pm$0.99} \\

    \hline
    \multirow{4}*{Average}
&79.30	 &81.84	&87.84	 &73.12	& 80.42	& 65.27 & 67.34 &	\underline{88.74}	& 84.90 &	\textbf{93.28}\\
&65.80	 &67.50	&\underline{82.32}	 &60.64	&67.36	 &44.04	&45.27 &	80.77 &	78.84 &	\textbf{87.42}\\
&56.48 	& 59.67	&71.78	&45.89	&57.51	&37.80&	39.15	&\underline{71.92}	&66.80	&\textbf{79.81}\\
&77.67 &	79.63 &	85.18	& 73.04 &	78.15 &	66.50	& 67.89 &	\underline{86.04} &	82.79	&\textbf{90.16}\\
     \hline
    \end{tabular*}
    }
\end{table*}

\section{Experiments} \label{sec4}
In the experimental section, we conducted a comprehensive evaluation of our proposed method RFLkPC, by comparing it with several existing models related to plane clustering. The selected models include KPC \cite{bradley2000k}, FkPC \cite{zhu2008}, LkPPC \cite{yang2015local}, SCC \cite{chen2009}, DPCP-KSS \cite{ding2021dual}, FCRM \cite{hathaway1993}, FWCHR \cite{zhao2018}, LFDC \cite{QI2023locally}, and SNMoE \cite{chamroukhi2016}. The model parameters for all experiments followed the researchers' recommendations. The parameters $c_1$ and $c_2$ in LkPPC are selected in $\{2^i|i=-8,-7,\dots,0\}$ and $\{2^j|j=-8,-7,\dots,7\}$. Parameter $\gamma$ of the FWCHR is selected from $\{0.5,2,10,100\}$.
Parameter $\mu$ of the LFDC is selected from $\{10^i|i=-8,-7,\dots,8\}$. The parameters $\alpha$ and $\lambda$ in our RFLkPC are selected from $\{0,0.1,0.2,\dots,1\}$ and $\{10^i|i=-4,-3,\dots,0\}$ respectively. Under the same initialization with Kmeans, the above parameters are selected from these sets with the 10-fold cross-validation.

{{To evaluate the performance of our method, we employed four commonly used clustering evaluation metrics, namely Accuracy (ACC) \cite{Chen2020}, Normalized Mutual Information (NMI) \cite{danon2005}, Adjusted Rand Index (ARI) \cite{steinley2004} and Purity \cite{amigo2009} to measure the effectiveness of the clustering results obtained by each model on both simulated data and real-world datasets. The results obtained from these experiments will shed light on the effectiveness and robustness of our approach in different scenarios. All experiments were carried out on a Windows 10 computer equipped with an Intel Core i5 processor and 8GB of RAM under the MATLAB 2019b platform.}}

\subsection{Simulated data} \label{sec4.1}
In order to reflect the advantages of boundedness and mixture distance, we construct the following synthetic datasets Si $(i=1,2,3)$ and its variants in Fig. \ref{fig-Si}, which contain three types of plane clustering scenarios with different noise or outliers. {{The experiments are run 100 times with Kmeans initialization, and the mean clustering scores of ACC, NMI, ARI and Purity are shown in Table \ref{Tab-Si-score}.} The best result for each metric is highlighted in bold, while the second-best is underlined. According to Table \ref{Tab-Si-score}, on most of the simulated data, RFLkPC has achieved the best or suboptimal results. Overall, RFLkPC performs the best among the methods, followed by LFDC and LkPPC.}

The dataset Si $(i=1,2,3)$ in the first row of Fig. \ref{fig-Si} does not contain any outliers. These datasets are utilized to examine the impact of local boundedness on the plane clusters. It is evident that the effect of local boundedness on S2 is significantly more pronounced compared to S1. When models lacking local bounded control are applied to S2, they encounter challenges in distinguishing between clusters 2 and 3 and their clustering score drops dramatically, such as KPC, FkPC, DPCP-KSS, FCRM and FWCHR. Furthermore, it can be observed that RFLkPC outperforms FkPC significantly, reinforcing the aforementioned point. The main difference between S3 and S1 is the presence of an additional cluster with overlapping in S3. Compared to S1, the increase in the number of clusters and the presence of overlap in S3 make clustering more difficult. As a result, the scores of all methods are significantly decreased. However, models that utilize robust distance such as LFDC and RFLkPC are still able to obtain acceptable results despite these challenges.

To investigate the impact of noise on plane clustering performance, we further simulate four types of noisy datasets: Gaussian-distributed noise (Si-N), Laplace-distributed noise (Si-L), t-distributed noise with one degree of freedom (Si-T), and uniformly distributed noise with 40\% random noise points added (Si-UN). As shown in Table~\ref{Tab-Si-score}, Gaussian noise has a relatively minor impact on all plane clustering algorithms, with Si-N showing only a slight drop in performance compared to the clean dataset (Si). In contrast, the heavy-tailed Laplace (Si-L) and t-distributed (Si-T) noise lead to a more substantial decline in clustering accuracy, indicating that heavy-tailed noise significantly increases the difficulty for clustering methods.

Among the compared methods, RFLkPC achieves the best performance on the Si-N and Si-T datasets, demonstrating its robustness to both light and heavy-tailed noise. On S1-L, DPCP-KSS achieves the highest score, followed closely by RFLkPC. On S2-L, RFLkPC again outperforms others, with LkPPC and SNMoE ranking next. For S3-L, KPC attains the best performance, followed by RFLkPC and LFDC.

For the Si-UN datasets, where 40\% uniformly distributed noise points are introduced, the clustering performance drops more significantly across all methods compared to the other noisy variants. However, this trend is not observed on S2-UN, where the added noise disrupts the local boundedness structure of the clusters. Interestingly, this disruption appears to aid certain methods in distinguishing clusters 2 and 3, possibly due to their overlapping orientations. Despite this, RFLkPC consistently achieves either the best or second-best performance across all Si-UN datasets. Specifically, on S1-UN, LFDC ranks first, followed by RFLkPC and SNMoE. On S2-UN, LkPPC performs best, with RFLkPC and SNMoE following closely. On S3-UN, RFLkPC outperforms all other methods, followed by FkPC and LkPPC.

\subsection{Real Data}

\begin{figure}[!t]
\centering
\begin{minipage}[b]{0.49\columnwidth}
  \centering
  \includegraphics[width=0.95\linewidth]{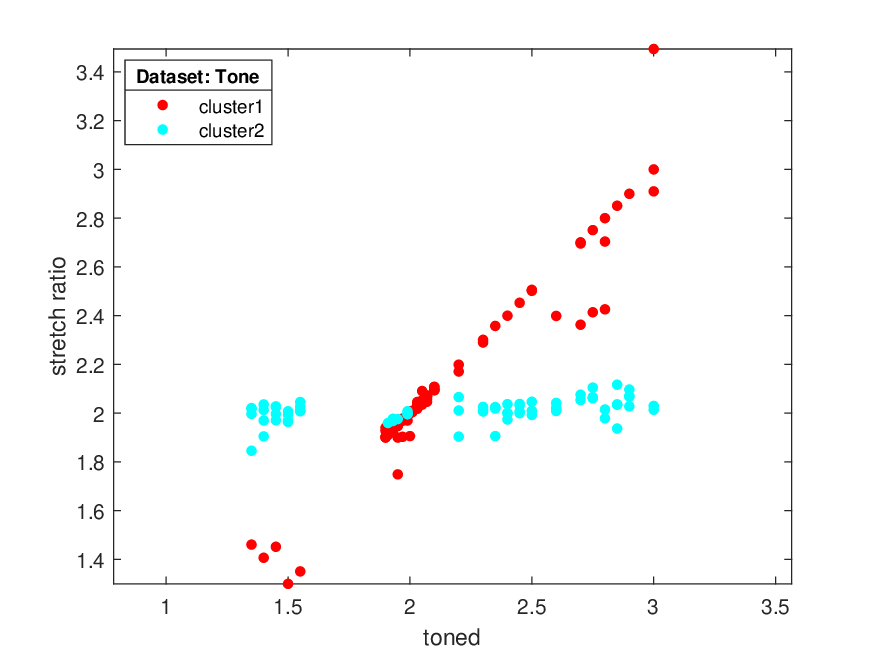}
  \caption{Tone perception data.}
  \label{fig-tone}
\end{minipage}
\hfill
\begin{minipage}[b]{0.49\columnwidth}
  \centering
  \includegraphics[width=0.95\linewidth]{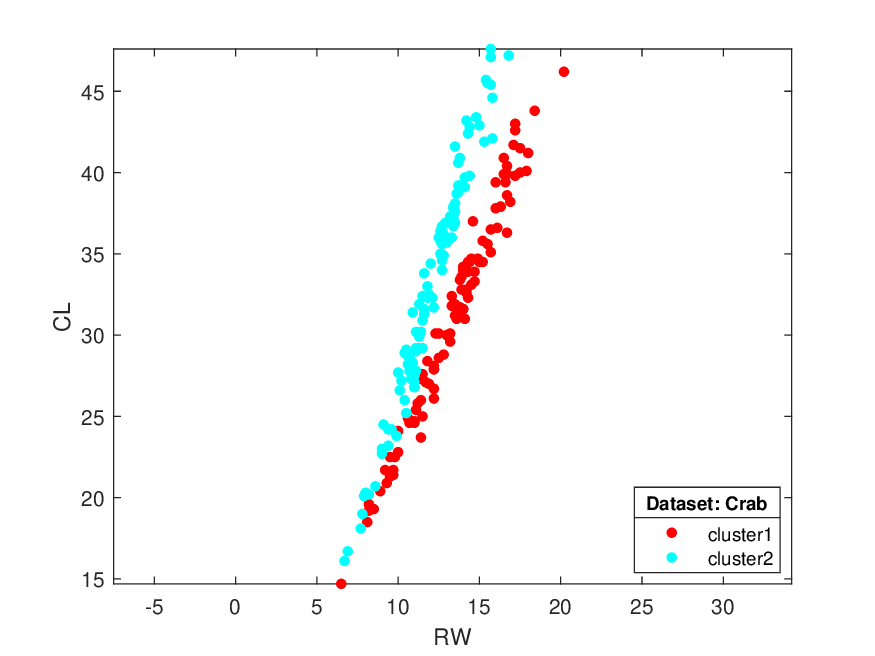}
  \caption{Crab data.}
  \label{fig-crab}
\end{minipage}
\end{figure}

\begin{table*}[h]
\centering
\caption{Mean score of  Various Methods on real data Tone and Crab.}
\label{Tab-tone-crab}
\resizebox{1.85\columnwidth}{!}{
\begin{tabular*}{1.05\linewidth}{>{}l >{}l >{}l >{}l >{}l >{}l >{}l >{}l >{}l >{}l >{}l}
    \hline
    Dataset & KPC & FkPC & LkPPC & SCC & DPCP-KSS & FCRM &FWCHR &LFDC &SNMoE &RFLkPC\\
    \hline
    \multirow{4}*{~} &ACC(\%) &ACC(\%) &ACC(\%) &ACC(\%) &ACC(\%) &ACC(\%) &ACC(\%) &ACC(\%) &ACC(\%) &ACC(\%)\\
    ~ &NMI(\%) &NMI(\%) &NMI(\%) &NMI(\%) &NMI(\%) &NMI(\%) &NMI(\%) &NMI(\%) &NMI(\%) &NMI(\%)\\
~ &ARI(\%) &ARI(\%) &ARI(\%) &ARI(\%) &ARI(\%) &ARI(\%) &ARI(\%) &ARI(\%) &ARI(\%) &ARI(\%)\\
~ &Purity(\%) &Purity(\%) &Purity(\%) &Purity(\%) &Purity(\%) &Purity(\%) &Purity(\%) &Purity (\%) &Purity(\%) &Purity(\%)\\
    \hline
& 87.07$\pm$1.13 & \underline{94.00$\pm$0.00} & 88.64$\pm$1.31 & 65.76$\pm$0.09 & 92.00$\pm$0.00 & 93.33$\pm$0.00 & \textbf{96.00$\pm$0.00} & 92.00$\pm$0.00 & 68.57$\pm$0.31 & \textbf{96.00$\pm$0.00} \\
Tone & 57.25$\pm$2.37 & \underline{72.92$\pm$0.00} & 60.15$\pm$2.74 & 7.38$\pm$0.07  & 59.82$\pm$0.00 & 70.91$\pm$0.00 & \textbf{76.95$\pm$0.00} & 67.33$\pm$1.17 & 27.53$\pm$0.36 & \textbf{76.95$\pm$0.00} \\
N(150) & 59.71$\pm$2.61 & \underline{77.29$\pm$0.00} & 65.13$\pm$2.86 & 9.15$\pm$0.10  & 70.36$\pm$0.00 & 74.94$\pm$0.00 & \textbf{84.54$\pm$0.00} & 70.36$\pm$0.00 & 12.59$\pm$0.46 & \textbf{84.54$\pm$0.00} \\
& 87.07$\pm$1.13 & \underline{94.00$\pm$0.00} & 88.64$\pm$1.31 & 65.76$\pm$0.09 & 92.00$\pm$0.00 & 93.33$\pm$0.00 & \textbf{96.00$\pm$0.00} & 92.00$\pm$0.00 & 68.57$\pm$0.31 & \textbf{96.00$\pm$0.00} \\
    \hline
&96.17$\pm$0.03 & \underline{96.50$\pm$0.00} & 96.00$\pm$0.00 & 91.80$\pm$0.27 & \textbf{97.00$\pm$0.00} & 95.50$\pm$0.00 & 95.00$\pm$0.00 & 95.94$\pm$0.07 & 94.50$\pm$0.00 & \underline{96.50$\pm$0.00} \\
Crab &77.19$\pm$0.14 & \underline{78.16$\pm$0.00} & 75.94$\pm$0.00 & 63.44$\pm$1.56 & \textbf{80.56$\pm$0.00} & 77.82$\pm$0.00 & 73.85$\pm$0.00 & 93.77$\pm$0.17 & 70.88$\pm$0.00 & \underline{78.16$\pm$0.00} \\
N(200) &85.65$\pm$0.01 & \underline{86.42$\pm$0.00} & 84.56$\pm$0.00 & 72.38$\pm$0.14 & \textbf{88.30$\pm$0.00} & 82.72$\pm$0.00 & 80.90$\pm$0.00 & 83.85$\pm$0.03 & 79.11$\pm$0.00 & \underline{86.42$\pm$0.00} \\
&96.17$\pm$0.03 & \underline{96.50$\pm$0.00} & 96.00$\pm$0.00 & 91.80$\pm$0.27 & \textbf{97.00$\pm$0.00} & 95.50$\pm$0.00 & 95.00$\pm$0.00 & 95.94$\pm$0.07 & 94.50$\pm$0.00 & \underline{96.50$\pm$0.00} \\
     \hline
 \multirow{2}*{Average}
     & 92.74 & 95.25 & 92.32 & 78.78 & 94.50 & 94.42 & \underline{95.50} & 94.46 & 81.54 & \textbf{96.25} \\
     ~   & 69.57 & \underline{75.75} & 68.05 & 35.41 & 70.70 & 74.37 & 75.40 & 72.02 & 48.42 & \textbf{78.22} \\
     & 72.68 & 81.86 & 74.85 & 40.77 & 79.33 & 78.83 & \underline{82.72} & 77.11 & 45.85 & \textbf{85.48} \\
     & 92.74 & 95.25 & 92.32 & 78.78 & 94.50 & 94.42 & \underline{95.50} & 94.46 & 81.54 & \textbf{96.25} \\
     \hline
    \end{tabular*}
    }
\end{table*}

{{In this subsection we further apply our proposed algorithm to several real-world examples, including the Tone data, the Crab data and other benchmark datasets obtained from the UCI machine learning repository (http://www.ics.uci.edu/ml/). Unless otherwise specified, all real-world data are preprocessed to scale features to [0,1] range using column-wise min-max normalization.}}

1) \textbf{Tone data}: The tone perception data stem from an experiment by Cohen \cite{cohen1984} and have been analyzed using different methods by many researchers. A pure fundamental tone is played to a trained musician, in which electronically generated overtones are added, determined by a stretching ratio (stretch ratio). The data analyzed here belong to 150 trials (tuned, stretch-ratio) with the same musician, which contains two plane clusters in Fig. \ref{fig-tone}, and its categories are manually labeled by ourselves.

2) \textbf{Crab data}: The crab data from Campbell \cite{campbell1974} describes 5 morphological measurements of  \textit{Leptograpsus} \textit{Variegatus} crabs collected at Fremantle (Australia). We can distinguish the sex of crabs by different morphological characteristics. In this example, we adopt the approach described in \cite{zhao2018} to select variables RW (rear width in mm), CL (carapace length in mm) and SEX as a latent variable, which serves as the data's label for clustering evaluation and unknown to clustering methods. The crab data is shown in Fig. \ref{fig-crab}.

\begin{table}[h]
\centering
\caption{Details of  Some UCI Datasets.}
\label{Tab-HDataset}
\resizebox{0.95\columnwidth}{!}{
    \begin{tabular}{c c c c c c c}
    \hline
    datasets &Wine &Dermatology &Ionosphere &Vehicle &BUPA &Statlog\\
    \hline
    $N$ (size) &178 &366 &351 &846 &345 &6435\\
    $D$ (dim) &13 &34 &34 &18 &5 &36\\
    $K$ (clusters) &3 &6 &2 &4 &3 &6\\
    \hline
    \end{tabular}
    }
\end{table}

\begin{figure}[!h] 
  \centering
  \includegraphics[width=0.88\columnwidth]{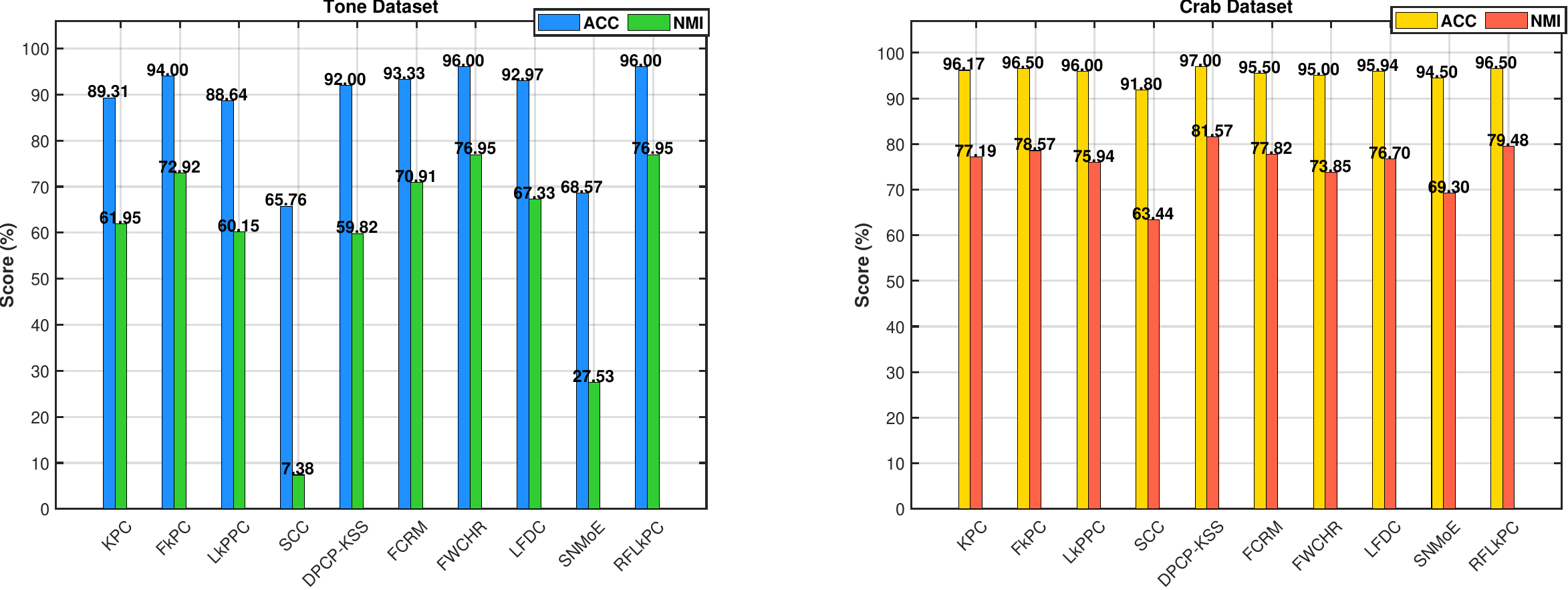}  
  \caption{{Results on Tone and Crab data for all competing methods.}} \label{resluts_tc}
\end{figure}

{{The experiment was run 100 times under the same Kmeans initialization, and the average performance of the Tone and Crab data on each method is show in Table \ref{Tab-tone-crab} and Figure \ref{resluts_tc}. It is worth noting that these two datasets do not exhibit locally bounded cluster structures; instead, they primarily serve to evaluate the robustness of clustering methods against noise.}}

On the Tone data, RFLkPC and FWCHR demonstrate the best performance, with FkPC following closely behind. On the Crab data, DPCP-KSS exhibits the highest performance, followed by RFLkPC and FkPC. In general, RFLkPC shows greater robustness, and each algorithm performs better on the Crab data compared to the Tone data, thanks to the good quality of the data. Additionally, due to the presence of some overlap between clusters in both Tone and Crab, the fuzzy clustering method, such as FkPC, outperforms the hard clustering method, such as KPC and LkPPC.

3) \textbf{Other benchmark datasets}: The data dimensions for the aforementioned simulated and real experiments were all 2-dimensional. Here, we further compare the performance of our method with relevant plane clustering algorithms on higher-dimensional real-world data. These datasets can be downloaded from the UCI Machine Learning Repository (http://www.ics.uci.edu/ml/), and their detailed information is provided in Table \ref{Tab-HDataset}.

It should be noted that the cluster structure of these data may not necessarily conform to a multi-plane cluster structure, and the response variables are also unknown. Therefore, some methods based on mixture regression are no longer compared. We primarily compare our RFLkPC with methods such as KPC \cite{bradley2000k}, FkPC \cite{zhu2008}, LkPPC \cite{yang2015local}, SCC \cite{chen2009}, DPCP-KSS \cite{ding2021dual}, and LFDC \cite{QI2023locally}, all of which require the estimation of plane normals. It is worth mentioning that for the LFDC \cite{QI2023locally}, we only compare it in its one-dimensional case, assuming that the cluster structure consists of multiple planes rather than flats.

\begin{table*}
\centering
\caption{{Mean score of  Various Plane Clustering Methods on real UCI datasets.}}
\label{Tab-UCI-data}
    \resizebox{1.58\columnwidth}{!}{
\begin{tabular*}{0.82\linewidth}{>{}l >{}l >{}l >{}l >{}l >{}l >{}l >{}l}
    \hline
    Dataset &KPC &FkPC &LkPPC &SCC &DPCP-KSS &LFDC &RFLkPC\\
    \hline
    \multirow{2}*{~} &ACC(\%) &ACC(\%) &ACC(\%) &ACC(\%) &ACC(\%) &ACC(\%) &ACC(\%)\\
    $N\times D\times K$ &NMI(\%) &NMI(\%) &NMI(\%) &NMI(\%) &NMI(\%) &NMI(\%) &NMI(\%)\\
~ &ARI(\%) &ARI(\%) &ARI(\%) &ARI(\%) &ARI(\%) &ARI(\%) &ARI(\%)\\
~ &Purity(\%) &Purity(\%) &Purity(\%) &Purity(\%) &Purity (\%) &Purity(\%) &Purity(\%)\\
    \hline
& 64.19$\pm$0.34 & 61.01$\pm$0.08 & \textbf{96.02$\pm$0.57} & 95.11$\pm$0.21 & 66.49$\pm$0.75 & 95.45$\pm$1.47 & \underline{95.84$\pm$0.03} \\
Wine & 33.93$\pm$0.39 & 30.60$\pm$0.05 & \textbf{87.45$\pm$0.52} & 83.90$\pm$0.57 & 35.96$\pm$0.89 & \underline{86.51$\pm$2.44} & 85.52$\pm$0.07 \\
$178\times13\times3$ &27.06$\pm$0.41 & 24.40$\pm$0.06 & \textbf{89.01$\pm$0.52} & 81.38$\pm$0.32 & 10.46$\pm$0.81 & 82.89$\pm$2.53 & \underline{84.83$\pm$0.00} \\
&64.19$\pm$0.34 & 61.01$\pm$0.08 & \textbf{96.02$\pm$0.57} & 95.11$\pm$0.21 & 66.49$\pm$0.75 & 95.45$\pm$1.47 & \underline{95.84$\pm$0.03} \\
    \hline
     &36.74$\pm$0.90 &34.98$\pm$0.71 &\underline{67.34$\pm$3.56} &22.68$\pm$0.00 &35.07$\pm$0.86 &32.04$\pm$0.49 &\textbf{74.73$\pm$0.40}\\
    Dermatology &28.18$\pm$1.55 &19.69 $\pm$1.20 &\underline{61.59$\pm$5.50} &1.95$\pm$0.00 &24.72$\pm$1.65 &6.96$\pm$1.83 &\textbf{76.33$\pm$0.46}\\
$366\times34\times6$ & 8.96$\pm$0.80 & 5.87$\pm$0.85 & \underline{54.40$\pm$5.34} & 1.95$\pm$0.00 & 2.54$\pm$2.16 & 4.13$\pm$1.90 & \textbf{70.34$\pm$0.60} \\
&36.74$\pm$0.90 &34.98$\pm$0.71 &\underline{67.34$\pm$3.56} &22.68$\pm$0.00 &35.07$\pm$0.86 &32.04$\pm$0.49 &\textbf{74.73$\pm$0.40}\\
     \hline
      &53.85$\pm$0.12 &55.56$\pm$0.00 &\textbf{71.23$\pm$0.00} &\underline{71.08$\pm$0.15} &52.28$\pm$0.06 &70.03$\pm$0.56 &\textbf{71.23$\pm$0.00}\\
    Ionosphere &0.27$\pm$0.03 &0.28$\pm$0.00 &\textbf{13.49$\pm$0.00} &\underline{13.10$\pm$0.18} &0.10$\pm$0.00 &11.84$\pm$0.78 &\textbf{13.49$\pm$0.00}\\
  $351\times18\times2$ & 0.35$\pm$0.05 & 0.70$\pm$0.00 & \textbf{17.76$\pm$0.00} & 17.66$\pm$0.31 & 1.93$\pm$0.24 & \underline{17.68$\pm$0.92} & \textbf{17.76$\pm$0.00} \\
& 64.1$\pm$0.00  & 64.1$\pm$0.00  & \textbf{71.23$\pm$0.00} & \underline{71.08$\pm$0.00} & 64.1$\pm$0.00  & 70.03$\pm$0.00 & \textbf{71.23$\pm$0.00} \\
     \hline
      &29.84$\pm$0.19 &29.60$\pm$0.20 &\textbf{36.78$\pm$0.24}  & 30.33$\pm$0.46 & 30.62$\pm$0.40 & 32.92$\pm$0.64 &\underline{36.18$\pm$0.16}\\
    Vehicle & 1.06$\pm$0.07 &1.03$\pm$0.07 &\underline{10.01$\pm$0.21} & 6.25$\pm$0.43  & 1.62$\pm$0.18  & 4.43$\pm$0.58 &\textbf{10.22$\pm$0.04}\\
$846\times18\times4$ & 1.03$\pm$0.21 & 0.61$\pm$0.09 & \textbf{4.72$\pm$0.22} & 0.83$\pm$0.81 & 1.16$\pm$0.19 & 2.64$\pm$0.36 & \underline{3.85$\pm$0.06} \\
&29.84$\pm$0.19 &29.60$\pm$0.20 &\textbf{36.78$\pm$0.24}  & 30.33$\pm$0.46 & 30.62$\pm$0.40 & 32.92$\pm$0.64 &\underline{36.18$\pm$0.16}\\
     \hline
&38.01$\pm$0.30 & 40.11$\pm$0.40 & 42.82$\pm$0.52 & 41.73$\pm$0.38 & 39.09$\pm$0.19 & \textbf{43.71$\pm$0.26} & \underline{43.19$\pm$0.00} \\
BUPA &1.38$\pm$0.12  & 2.01$\pm$0.18  & 3.52$\pm$0.26  & 3.34$\pm$0.21  & 1.98$\pm$0.12  & \textbf{4.96$\pm$0.22}  & \underline{4.29$\pm$0.01}  \\
$345\times5\times3$ &0.56$\pm$0.09  & 1.21$\pm$0.15  & 2.37$\pm$0.20  & 3.10$\pm$0.23  & 1.39$\pm$0.11  & \textbf{4.79$\pm$0.23}  & \underline{4.24$\pm$0.00}  \\
&38.01$\pm$0.30 & 40.11$\pm$0.40 & 42.82$\pm$0.52 & 41.73$\pm$0.38 & 39.09$\pm$0.19 & \textbf{43.71$\pm$0.26} & \underline{43.19$\pm$0.00} \\
     \hline            
     ~ &19.86$\pm$0.00 &19.68$\pm$0.00 &\underline{66.70$\pm$0.00} &27.28$\pm$0.00 &23.20$\pm$0.00 &23.47$\pm$0.01 &\textbf{70.91$\pm$0.00}\\
      Statlog(Landsat) &0.83$\pm$0.00 &0.79$\pm$0.00 &\textbf{61.49$\pm$0.00} &17.76$\pm$0.00 &3.40$\pm$0.00 &3.62$\pm$0.01 &\underline{60.21$\pm$0.00}\\
    $4435\times36\times6$ &0.35$\pm$0.00 &0.30$\pm$0.00 &\underline{53.47$\pm$0.00} &5.35$\pm$0.00 &0.98$\pm$0.00 &1.88$\pm$0.01 &\textbf{53.88$\pm$0.00}\\            
 ~ &25.37$\pm$0.00 &24.33$\pm$0.00 &\textbf{75.13$\pm$0.00} &32.72$\pm$0.00 &27.17$\pm$0.00 &28.05$\pm$0.01 &\underline{74.25$\pm$0.00}\\  
     \hline
    \multirow{4}*{Average}
     & 40.42 & 40.16 & \underline{63.48} & 48.04 & 41.13 & 49.60 & \textbf{65.35} \\
     ~  &10.94 & 9.07 & \underline{39.59} & 21.05 & 11.30 & 19.72 & \textbf{41.68} \\
     &6.39 & 5.52 & \underline{36.96} & 18.38 & 3.08 & 18.83 & \textbf{39.15} \\
     ~  &43.04 & 42.36 & \underline{64.89} & 48.94 & 43.76 & 50.37 & \textbf{65.90} \\
     \hline
    \end{tabular*}
    }
\end{table*}

\begin{figure}[!h] 
  \centering
  \includegraphics[width=0.93\columnwidth]{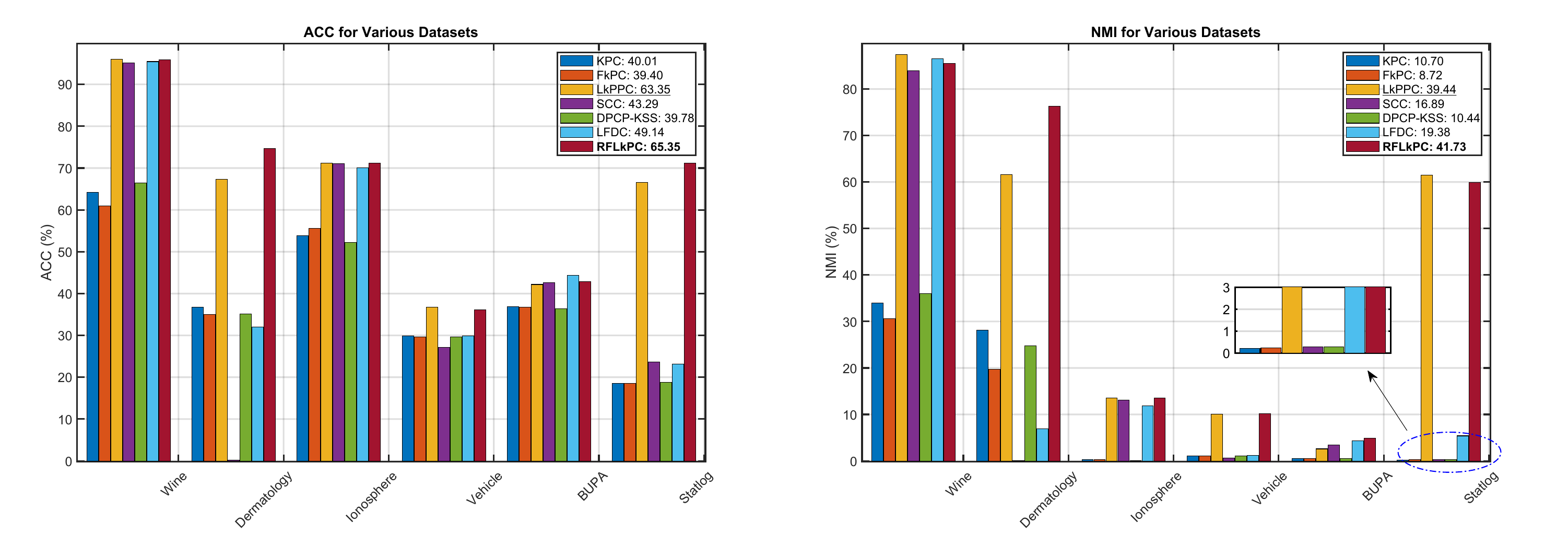}  
  \caption{{Results on UCI datasets for all competing methods.}} \label{resluts_uci}
\end{figure}

{{Furthermore, due to the presence of irrelevant features and redundancy in the Ionosphere dataset, we apply PCA \cite{wold1987PCA} for dimensionality reduction reducing the dimensionality from 34 to 15 while retaining 90\% of the original variance. Notably, unlike other real-world datasets in our experiments, the Ionosphere dataset is not preprocessed using min-max normalization. For the Statlog (Landsat) dataset, considering the computational time cost, we only utilize its training set, which consists of 4435 samples. All experimental results are presented in Table \ref{Tab-UCI-data} and Figure \ref{resluts_uci}.}}

{Overall, from Table IV, RFLkPC achieved the best or second-best scores on the majority of UCI datasets, closely followed by LkPPC and then LFDC. RFLkPC demonstrates a significant performance advantage over competing methods on both the Dermatology and Statlog (Landsat) datasets, with particularly pronounced gains observed on the Dermatology dataset. This can be attributed to several distinctive characteristics of the Dermatology dataset: (i) most features are discrete; (ii) there exists significant class imbalance; and (iii) although class labels are available, inter-class boundaries are not well-separated. Property (i) limits the applicability of conventional continuous-space distance metrics (e.g., Euclidean distance), which are ill-suited for discrete-valued features. Properties (ii) and (iii) increase the difficulty of clustering due to the uneven distribution of samples across six classes and the semantic overlap between categories, which leads to entangled feature distributions. These challenges necessitate the use of boundedness-aware constraints and robust distance formulations. The proposed method explicitly accounts for local boundedness and incorporates a more resilient similarity measure, making it particularly effective in handling such structural complexities, thereby yielding superior performance.}

In the context of plane clustering scenarios, LFDC was only considered in its one-dimensional case, which may limit its ability to perform multidimensional clustering. As a result, LFDC achieved the best performance only on the BUPA dataset, ranking third on the Wine and Vehicle datasets. On the other hand, for the high-dimensional subspace clustering models DPCP-KSS and SCC, despite their success in certain computer vision tasks, they heavily depend on effective initialization strategies and may not perform optimally in this particular context. Lastly, and more importantly, the RFLkPC model significantly improves clustering performance on all datasets by introducing the boundedness of plane clustering and robust distance, compared to the FkPC model. This indicates that the introduction of boundedness and robustness has successfully achieved our original intention of improving the clustering effectiveness of FkPC.

\subsection{{Point cloud data}}
\begin{figure}[!t] 
  \centering
  \includegraphics[width=0.56\columnwidth]{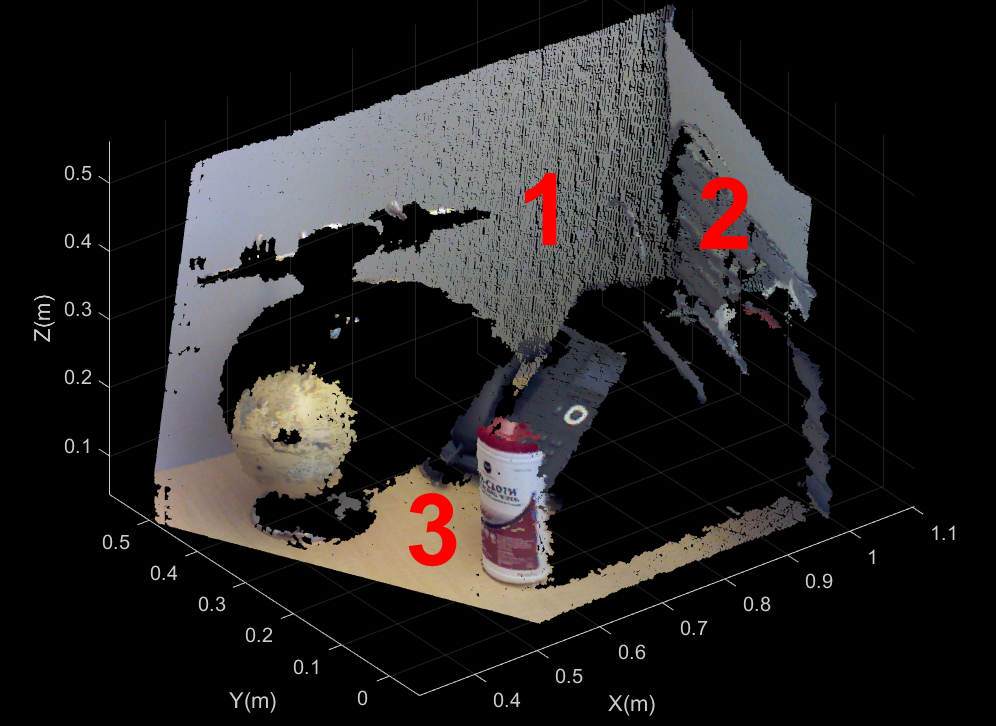}
  \caption{{The point cloud data contains 3 planes.}}
  \label{fig-pd}
\end{figure}

{In computer vision applications such as motion segmentation and robot path planning, it is often necessary to pre-identify structural elements in indoor scenes, such as walls, floors, and desktops. This task can be formulated as a plane detection or plane clustering problem. In this section, we demonstrate the applicability of the proposed RFLkPC method to 3D point cloud data, using a built-in dataset from MATLAB. The point cloud is captured using a Kinect depth sensor. Fig.~\ref{fig-pd} shows the raw point cloud and the resulting plane clusters, which include two walls, a desktop, and various miscellaneous objects (treated as outliers).}

\begin{figure}[!t] 
  \centering
  \includegraphics[width=1.0\columnwidth]{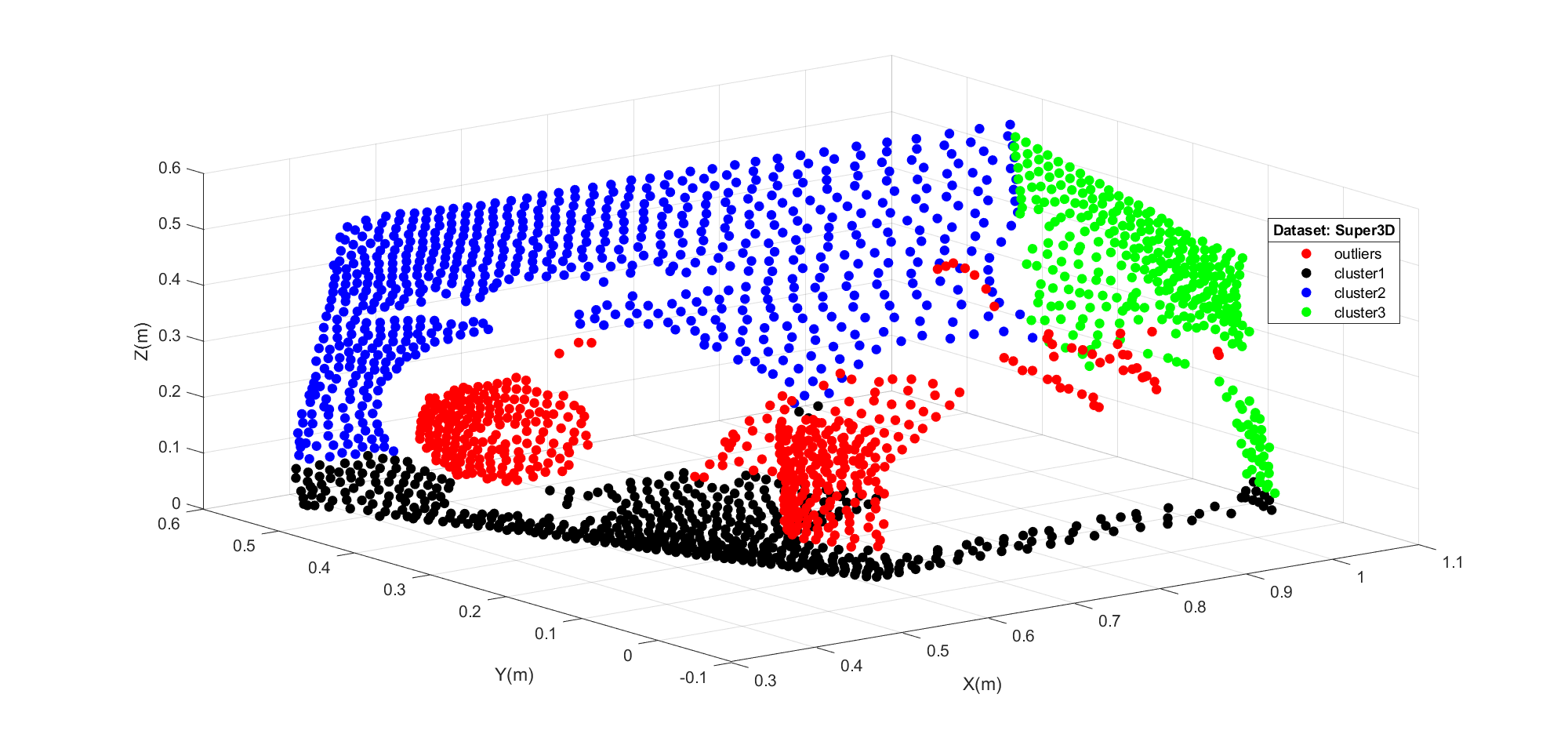}
  \caption{{The superpixel 3D points.}}
  \label{fig-super3d}
\end{figure}

Since the point cloud's depth data size is $480\times640\times3$, it initially contains 307,200 3D points. After removing some invalid 3D points, 204,728 valid samples remain. To further reduce the computational burden, we perform plane clustering on its superpixel-based representation, Super3D, which is constructed by computing the median of neighboring pixels without min-max normalization. This approach yields approximately 2000 superpixel 3D points, and their corresponding plane category information are depicted in Fig.\ref{fig-super3d}.

{As the 3D point cloud data contains vertical planes, certain regression coefficient estimation algorithms like FCRM or SNMoE are not applicable. Hence, we focus on evaluating the performance of plane clustering models, and the results are presented in Table \ref{Tab-super3d}.}

\begin{table}[!t]
\centering
\caption{{Mean score of  plane clustering Methods on the point cloud data.}}
\label{Tab-super3d}
    \resizebox{0.96\columnwidth}{!}{
    \begin{tabular}{>{}l >{}l >{}l >{}l >{}l >{}l >{}l }
    \hline
    Dataset &KPC &FkPC &LkPPC &SCC &DPCP-KSS &RFLkPC\\
    \hline
    \multirow{4}*{~} &ACC(\%) &ACC(\%) &ACC(\%) &ACC(\%) &ACC(\%) &ACC(\%)\\
    ~ &NMI(\%) &NMI(\%) &NMI(\%) &NMI(\%) &NMI(\%) &NMI(\%) \\
~ &ARI(\%) &ARI(\%) &ARI(\%) &ARI(\%) &ARI(\%) &ARI(\%) \\
~ &Purity(\%) &Purity(\%) &Purity(\%) &Purity(\%) &Purity (\%) &Purity(\%) \\
    \hline
& 72.57$\pm$1.85 & 72.48$\pm$1.88 & \underline{94.24$\pm$0.00} & 92.26$\pm$0.41 & 73.12$\pm$1.94  & \textbf{95.23$\pm$0.05} \\
Super3D & 43.91$\pm$2.70 & 43.45$\pm$2.76 & \underline{79.62$\pm$0.00} & 75.39$\pm$0.64 & 44.91$\pm$3.06  & \textbf{83.42$\pm$0.19} \\
N(1980) & 38.04$\pm$2.63 & 37.89$\pm$2.68 & \underline{61.08$\pm$0.00} & 60.08$\pm$0.78 & 39.10$\pm$2.80  & \textbf{63.79$\pm$0.13} \\
& 71.71$\pm$2.01 & 71.73$\pm$1.99 & \underline{92.37$\pm$0.00} & 90.90$\pm$0.42 & 72.12$\pm$2.03 & \textbf{93.15$\pm$0.12} \\
    \hline
    \end{tabular}
    }
\end{table}

{On the Super3D dataset, the best performing method is RFLkPC, followed by LkPPC and SCC. In this indoor scene, RFLkPC outperforms other models likely because it handles both inter-cluster boundedness constraints and the influence of outliers more effectively. Although LkPPC also incorporates local boundedness constraints, its design places more emphasis on inter-cluster separation to enhance robustness to outliers. As a result, in scenes where planar structures are distinctly separated, LkPPC achieves competitive results. Furthermore, an in-depth exploration of clusters' boundedness and separability is beyond the scope of this paper, but it can serve as a valuable direction for further research.}

\subsection{The Sensitivity of regularization parameters on RFLkPC model}
\begin{figure}[!h] 
  \centering
  \includegraphics[width=0.93\columnwidth]{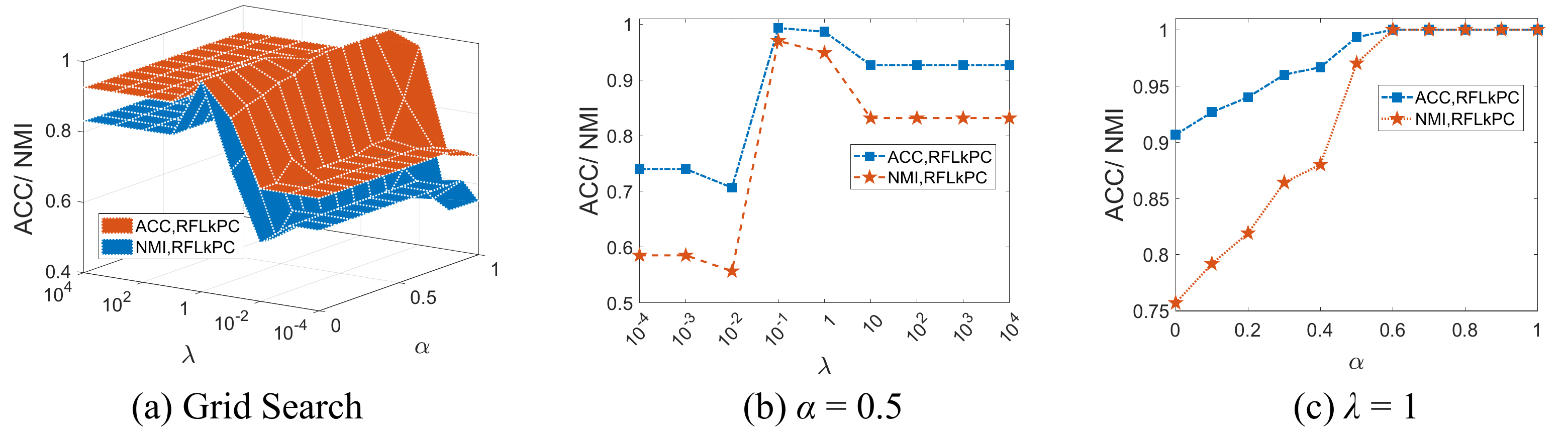}  
  \caption{Sensitivity of RFLkPC model to different regularization parameters.} \label{Parameters}
\end{figure}

The RFLkPC model fundamentally modifies the original objective function by incorporating regularization penalties. As shown in Equation (\ref{RFLkPC}), the model introduces two additional parameters: $\alpha$ and $\lambda$. Since the RFLkPC model involves two regularization parameters, a grid search method can be employed to select these parameters and further analyze their impact on the model's performance. The representative synthetic dataset S2 (see Figure \ref{fig-Si}, Section \ref{sec4.1}) is selected as the test function, and the grid search results are shown in Figure \ref{Parameters}(a). The parameter $\alpha$ follows an arithmetic progression with a range of $[0,1]$ and a common difference of 0.1; while $\lambda$ follows a geometric progression with a range of $[10^{-4}, 10^4]$ and a common ratio of $10^1$.

From Figure \ref{Parameters}(a), it is evident that parameters $\alpha$ and $\lambda$ have a markedly different impact on model accuracy. Specifically, when alpha is fixed, lambda has a more significant impact on the model's accuracy. In contrast, when lambda is fixed, changes in alpha have a relatively smaller effect. For example, when $\alpha=0.5$ or $\lambda=1$ is fixed, the effects of parameter $\lambda$ and parameter $\alpha$ on the accuracy of the RFLkPC model are recorded in Figures \ref{Parameters}(b) and \ref{Parameters}(c), respectively. Therefore, for the RFLkPC model, the regularization parameter $\lambda$ is more sensitive than the weight coefficient $\alpha$, particularly across different magnitudes. This sensitivity suggests that when adjusting the model parameters, prioritizing the tuning of $\lambda$ is more effective. Once an appropriate $\lambda$ is identified, different ranges of $\alpha$ can then be explored. Such an operation can not only obtain more accurate regularization parameters but also improve the efficiency of the parameter search process.

\subsection{Comparison of computational complexity and CPU time}

{To provide deeper insights into computational scalability, Table~\ref{Computational complexity} summarizes the theoretical complexities of representative plane clustering methods. Classical approaches such as KPC and SCC exhibit polynomial time complexity with respect to the number of samples $N$, feature dimension $D$, and cluster count $K$, with SCC incurring additional overhead due to subspace estimation. In contrast, the proposed RFLkPC achieves a complexity of $\mathcal{O}(TNKD^3)$, where $T$ denotes the number of iterations. Despite the cubic dependency on $D$, the overall runtime remains tractable owing to the method’s rapid convergence and reduced iteration count.}

{Furthermore, to evaluate actual computational efficiency, we benchmarked the CPU runtime of each method on several real-world datasets from the UCI repository, as shown in Table~\ref{Time-UCI}. For smaller datasets (e.g., \textit{Wine}, \textit{BUPA}), all methods exhibit negligible execution times. As dimensionality and sample size increase, runtime disparities become more evident. SCC suffers from substantial computational burden—especially on the \textit{Statlog} dataset—largely due to costly spectral steps and convergence instability in complex scenarios. In comparison, RFLkPC consistently demonstrates favorable efficiency, offering significant speedups over SCC while remaining competitive with lightweight baselines.}

{Overall, the proposed RFLkPC strikes a desirable balance between theoretical scalability and practical efficiency, making it suitable for large-scale clustering tasks involving high-dimensional and noisy data.}

\begin{table}
\centering
\caption{{Computational complexity of Various Plane Clustering Methods.}}
\label{Computational complexity}
{
\resizebox{0.86\columnwidth}{!}{
\begin{tabular*}{0.95\linewidth}{ccc}
\hline
Methods & Computational complexity & Bounded constraint\\
\hline
KPC \cite{bradley2000k} & $\mathcal{O}(TN(KD+D^2))$ & \ding{55} \\
FkPC \cite{zhu2008} & $\mathcal{O}(TN(KD^2+NK^2))$ & \ding{55} \\
LkPPC \cite{yang2015local} & $\mathcal{O}(TNKD^2)$ & \ding{51} \\
SCC \cite{chen2009} & $\mathcal{O}(TNKD^3)$ & \ding{55} \\
DPCP-KSS \cite{ding2021dual} & $\mathcal{O}(TNKD^2)$ & \ding{55} \\
FCRM \cite{hathaway1993} & $\mathcal{O}(TN(KD^2+K^2))$ & \ding{55} \\
FWCHR \cite{zhao2018} & $\mathcal{O}(TN(KD^2+K^2))$ & \ding{55} \\
LFDC \cite{QI2023locally} & $\mathcal{O}(TK(N^2D+D^3))$ & \ding{51} \\
SNMoE \cite{chamroukhi2016} & $\mathcal{O}(TNKD^2)$ & \ding{55}\\
RFLkPC (Ours) & $\mathcal{O}(TNKD^3)$ & \ding{51} \\
\hline
\end{tabular*}}
}
\end{table}

\begin{table}
\centering
\caption{{CPU Time (s) of Various Plane Clustering Methods on real UCI datasets.}}
\label{Time-UCI}
{
    \resizebox{0.95\columnwidth}{!}{
    \begin{tabular*}{1.53\linewidth}{ccccccccc}
    \hline
    Dataset & Size ($N \times D$) &KPC &FkPC &LkPPC &SCC &DPCP-KSS &LFDC &RFLkPC\\
    \hline
     Wine & $178\times13$ & $0.0018$ &	$0.0371$	& $0.0008$	& $0.4202$	& $0.0348$	& $0.0867$	& $0.0631$ \\

BUPA & $345\times5$ &0.0025 & 0.0720 & 0.0023 & 0.1776 & 0.0226 & 0.1133 & 0.1184 \\

Ionosphere & $351\times18$ &0.0008 & 0.0009 & 0.0024 & 2.0837 & 0.1510 & 0.0023 & 0.0080 \\

Dermatology & $366\times34$ &0.0012 & 0.0034 & 0.0081 & 6.3265 & 0.1921 & 0.0349 & 0.2782 \\

Vehicle & $846\times18$ &0.0112 & 0.2038 & 0.0200 & 2.1720 & 0.1300 & 0.0488 & 0.5972 \\
                       
Statlog & $4435\times36$ & 1.4312 & 10.5729 & 0.2214 & 83.3161 & 0.7414 & 16.0614 & 15.4164 \\
     \hline
    \end{tabular*}
    }
}
\end{table}

\section{Conclusion and Future Work} \label{sec5}
{This paper presents a novel robust plane clustering model, RFLkPC, grounded in the fuzzy clustering framework. To address the challenges posed by the unbounded expansion of plane clusters and the sensitivity of clustering methods to noise and outliers, RFLkPC incorporates a cluster-boundedness regularization term and a hybrid distance metric based on hinge loss and the $L_1$ norm, thereby improving robustness and flexibility. As plane-based clustering spans multiple research domains—such as switching regression, mixture regression, and fuzzy c-regression—the terminology and problem definitions are often inconsistent, which hinders cross-disciplinary communication and further advancement. To bridge this gap, we conducted a comprehensive review of relevant literature, hoping to facilitate better communication and development on this topic.}

{Through extensive experiments on synthetic and real-world datasets, including tone or crab recognition, UCI data, and point cloud data, RFLkPC demonstrates superior performance over existing approaches, underscoring its broad application potential. These diverse tasks reflect varying assumptions regarding boundedness, inter-cluster separability, and robustness, highlighting the need for flexible models that adapt to different real-world scenarios. Despite these promising results, RFLkPC still has several limitations that warrant further research:}

\begin{itemize}
    \item {\textbf{Scalability}: While effective on moderate-sized datasets, RFLkPC's computational cost increases significantly with data dimensionality and volume, due in part to the use of SVD. Future work could explore approximation techniques or optimization strategies to enhance scalability in high-resolution or large-scale applications.}
    
    \item {\textbf{Parameter Adjustment}: The model involves regularization parameters that may require careful tuning for different datasets or noise conditions. Incorporating adaptive or data-driven parameter selection methods, such as Bayesian optimization or meta-learning, could improve ease of use and robustness.}
    
    \item {\textbf{Boundedness-Separability}: Although RFLkPC emphasizes local boundedness and achieves superior accuracy, methods like LkPPC, which prioritize inter-cluster separability, exhibit faster computation while still maintaining competitive accuracy. Future we could focus on developing a unified framework that balances boundedness and separability adaptively.}
\end{itemize}

{Thereby, it can be further investigated the theoretical interplay between boundedness and separability, and how these assumptions influence clustering performance under different conditions. Lastly, given that plane clustering can be viewed as a special case of $k$-subspace or $k$-manifold clustering, it would be worthwhile to extend RFLkPC to handle low-dimensional subspace clustering problems, thereby broadening its applicability to other domains such as computer vision.}


\bibliographystyle{IEEEtran}
\bibliography{Reference}

\begin{thebibliography}{10}
\providecommand{\url}[1]{#1}
\csname url@samestyle\endcsname
\providecommand{\newblock}{\relax}
\providecommand{\bibinfo}[2]{#2}
\providecommand{\BIBentrySTDinterwordspacing}{\spaceskip=0pt\relax}
\providecommand{\BIBentryALTinterwordstretchfactor}{4}
\providecommand{\BIBentryALTinterwordspacing}{\spaceskip=\fontdimen2\font plus
\BIBentryALTinterwordstretchfactor\fontdimen3\font minus
  \fontdimen4\font\relax}
\providecommand{\BIBforeignlanguage}[2]{{%
\expandafter\ifx\csname l@#1\endcsname\relax
\typeout{** WARNING: IEEEtran.bst: No hyphenation pattern has been}%
\typeout{** loaded for the language `#1'. Using the pattern for}%
\typeout{** the default language instead.}%
\else
\language=\csname l@#1\endcsname
\fi
#2}}
\providecommand{\BIBdecl}{\relax}
\BIBdecl

\bibitem{MacQuuen1967}
J.~MacQuuen, ``Some methods for classification and analysis of multivariate
  observation,'' in \emph{Proceedings of the 5th Berkley Symposium on
  Mathematical Statistics and Probability}, 1967, pp. 281--297.

\bibitem{Bezdek1994}
J.~C. Bezdek and S.~K. Pal, ``Fuzzy models for pattern recognition,'' 1994.

\bibitem{Goldfeld1973}
S.~Goldfeld and R.~Quandt, ``The estimation of structural shifts by switching
  regressions,'' in \emph{Annals of Economic and Social Measurement, Volume 2,
  number 4}.\hskip 1em plus 0.5em minus 0.4em\relax NBER, 1973, pp. 475--485.

\bibitem{huang2017regression}
H.~Huang, ``Regression in heterogeneous problems,'' \emph{Statistica Sinica},
  pp. 71--88, 2017.

\bibitem{song2014}
W.~Song, W.~Yao, and Y.~Xing, ``Robust mixture regression model fitting by
  laplace distribution,'' \emph{Computational Statistics \& Data Analysis},
  vol.~71, pp. 128--137, 2014.

\bibitem{shan2021}
A.~Shan and F.~Yang, ``Bayesian inference for finite mixture regression model
  based on non-iterative algorithm,'' \emph{Mathematics}, vol.~9, no.~6, p.
  590, 2021.

\bibitem{spath1979}
H.~Sp{\"a}th, ``Algorithmus 39 classenweise lineare regression,''
  \emph{Computing}, vol.~22, pp. 367--373, 1979.

\bibitem{schlittgen2011}
R.~Schlittgen, ``A weighted least-squares approach to clusterwise regression,''
  \emph{AStA Advances in Statistical Analysis}, vol.~95, no.~2, pp. 205--217,
  2011.

\bibitem{tsakiris2017}
M.~C. Tsakiris and R.~Vidal, ``Hyperplane clustering via dual principal
  component pursuit,'' in \emph{International conference on machine
  learning}.\hskip 1em plus 0.5em minus 0.4em\relax PMLR, 2017, pp. 3472--3481.

\bibitem{ding2021dual}
T.~Ding, Z.~Zhu, M.~Tsakiris, R.~Vidal, and D.~Robinson, ``Dual principal
  component pursuit for learning a union of hyperplanes: Theory and
  algorithms,'' in \emph{International Conference on Artificial Intelligence
  and Statistics}.\hskip 1em plus 0.5em minus 0.4em\relax PMLR, 2021, pp.
  2944--2952.

\bibitem{tron2007}
R.~Tron and R.~Vidal, ``A benchmark for the comparison of 3-d motion
  segmentation algorithms,'' in \emph{2007 IEEE conference on computer vision
  and pattern recognition}.\hskip 1em plus 0.5em minus 0.4em\relax IEEE, 2007,
  pp. 1--8.

\bibitem{chen2009}
G.~Chen and G.~Lerman, ``Spectral curvature clustering (scc),''
  \emph{International Journal of Computer Vision}, vol.~81, pp. 317--330, 2009.

\bibitem{bako2011}
L.~Bako, ``Identification of switched linear systems via sparse optimization,''
  \emph{Automatica}, vol.~47, no.~4, pp. 668--677, 2011.

\bibitem{blavzivc2020}
S.~Bla{\v{z}}i{\v{c}} and I.~{\v{S}}krjanc, ``Hybrid system identification by
  incremental fuzzy c-regression clustering,'' in \emph{2020 IEEE International
  Conference on Fuzzy Systems (FUZZ-IEEE)}.\hskip 1em plus 0.5em minus
  0.4em\relax IEEE, 2020, pp. 1--7.

\bibitem{yang2008}
M.-S. Yang, K.-L. Wu, J.-N. Hsieh, and J.~Yu, ``Alpha-cut implemented fuzzy
  clustering algorithms and switching regressions,'' \emph{IEEE Transactions on
  Systems, Man, and Cybernetics, Part B (Cybernetics)}, vol.~38, no.~3, pp.
  588--603, 2008.

\bibitem{leski2015}
J.~M. Leski and M.~Kotas, ``On robust fuzzy c-regression models,'' \emph{Fuzzy
  Sets and Systems}, vol. 279, pp. 112--129, 2015.

\bibitem{nie2020}
F.~Nie, X.~Zhao, R.~Wang, X.~Li, and Z.~Li, ``Fuzzy k-means clustering with
  discriminative embedding,'' \emph{IEEE Transactions on Knowledge and Data
  Engineering}, vol.~34, no.~3, pp. 1221--1230, 2020.

\bibitem{blavzivc2019}
S.~Bla{\v{z}}i{\v{c}} and I.~{\v{S}}krjanc, ``Incremental fuzzy c-regression
  clustering from streaming data for local-model-network identification,''
  \emph{IEEE transactions on fuzzy systems}, vol.~28, no.~4, pp. 758--767,
  2019.

\bibitem{zhao2021}
X.~Zhao, F.~Nie, R.~Wang, and X.~Li, ``Robust fuzzy k-means clustering with
  shrunk patterns learning,'' \emph{IEEE Transactions on Knowledge and Data
  Engineering}, vol.~35, no.~3, pp. 3001--3013, 2021.

\bibitem{hu2022}
X.~Hu, X.~Liu, W.~Pedrycz, Q.~Liao, Y.~Shen, Y.~Li, and S.~Wang, ``Multi-view
  fuzzy classification with subspace clustering and information granules,''
  \emph{IEEE Transactions on Knowledge and Data Engineering}, vol.~35, no.~11,
  pp. 11\,642--11\,655, 2022.

\bibitem{leski2023}
J.~M. Leski, ``Fuzzy double-ordered c-regression models based on fuzzy
  s-estimators,'' \emph{Fuzzy Sets and Systems}, p. 108531, 2023.

\bibitem{bradley2000k}
P.~S. Bradley and O.~L. Mangasarian, ``K-plane clustering,'' \emph{Journal of
  Global optimization}, vol.~16, pp. 23--32, 2000.

\bibitem{shao2013}
Y.-H. Shao, L.~Bai, Z.~Wang, X.-Y. Hua, and N.-Y. Deng, ``Proximal plane
  clustering via eigenvalues,'' \emph{Procedia Computer Science}, vol.~17, pp.
  41--47, 2013.

\bibitem{yang2015local}
Z.-M. Yang, Y.-R. Guo, C.-N. Li, and Y.-H. Shao, ``Local k-proximal plane
  clustering,'' \emph{Neural Computing and Applications}, vol.~26, pp.
  199--211, 2015.

\bibitem{wang2015twin}
Z.~Wang, Y.-H. Shao, L.~Bai, and N.-Y. Deng, ``Twin support vector machine for
  clustering,'' \emph{IEEE transactions on neural networks and learning
  systems}, vol.~26, no.~10, pp. 2583--2588, 2015.

\bibitem{ye2017l1}
Q.~Ye, H.~Zhao, Z.~Li, X.~Yang, S.~Gao, T.~Yin, and N.~Ye, ``L1-norm distance
  minimization-based fast robust twin support vector $ k $-plane clustering,''
  \emph{IEEE transactions on neural networks and learning systems}, vol.~29,
  no.~9, pp. 4494--4503, 2018.

\bibitem{bai2019clustering}
L.~Bai, Y.-H. Shao, Z.~Wang, and C.-N. Li, ``Clustering by twin support vector
  machine and least square twin support vector classifier with uniform output
  coding,'' \emph{Knowledge-Based Systems}, vol. 163, pp. 227--240, 2019.

\bibitem{richhariya2020}
B.~Richhariya, M.~Tanveer, A.~D.~N. Initiative \emph{et~al.}, ``Least squares
  projection twin support vector clustering (lsptsvc),'' \emph{Information
  Sciences}, vol. 533, pp. 1--23, 2020.

\bibitem{tanveer2021sparse}
M.~Tanveer, M.~Tabish, and J.~Jangir, ``Sparse pinball twin bounded support
  vector clustering,'' \emph{IEEE Transactions on Computational Social
  Systems}, vol.~9, no.~6, pp. 1820--1829, Dec 2022.

\bibitem{tseng2000}
P.~Tseng, ``Nearest q-flat to m points,'' \emph{Journal of Optimization Theory
  and Applications}, vol. 105, pp. 249--252, 2000.

\bibitem{wang2011}
Y.~Wang, Y.~Jiang, Y.~Wu, and Z.-H. Zhou, ``Localized k-flats,'' in
  \emph{Proceedings of the AAAI Conference on Artificial Intelligence},
  vol.~25, 2011, pp. 525--530.

\bibitem{zhu2008}
L.~Zhu, S.-t. Wang, Y.-h. Pan, and B.~Han, ``Improved fuzzy partitions for
  k-plane clustering algorithm and its robustness research,'' \emph{Journal of
  Electronics and Information Technology}, vol.~30, no.~8, pp. 1923--1927,
  2008.

\bibitem{gu2020fuzzy}
S.~Gu, Y.~Nojima, H.~Ishibuchi, and S.~Wang, ``Fuzzy style k-plane
  clustering,'' \emph{IEEE Transactions on Fuzzy Systems}, vol.~29, no.~6, pp.
  1518--1532, 2020.

\bibitem{kumar2022}
P.~Kumar, D.~Kumar, and R.~K. Agrawal, ``Fuzzy k-plane clustering method with
  local spatial information for segmentation of human brain mri image,''
  \emph{Neural Computing and Applications}, pp. 1--20, 2022.

\bibitem{huang2013support}
X.~Huang, L.~Shi, and J.~A. Suykens, ``Support vector machine classifier with
  pinball loss,'' \emph{IEEE transactions on pattern analysis and machine
  intelligence}, vol.~36, no.~5, pp. 984--997, 2014.

\bibitem{QI2023locally}
Y.-F. Qi, Y.-H. Shao, C.-N. Li, and Y.-R. Guo, ``Locally finite distance
  clustering with discriminative information,'' \emph{Information Sciences},
  vol. 623, pp. 607--632, 2023.

\bibitem{zou2005regularization}
H.~Zou and T.~Hastie, ``Regularization and variable selection via the elastic
  net,'' \emph{Journal of the Royal Statistical Society Series B: Statistical
  Methodology}, vol.~67, no.~2, pp. 301--320, 2005.

\bibitem{hosmer1978}
D.~W. Hosmer, ``Estimating mixtures of normal distributions and switching
  regressions: Comment,'' \emph{Journal of the American statistical
  association}, vol.~73, no. 364, pp. 741--744, 1978.

\bibitem{Jain2002}
M.~Figueiredo and A.~Jain, ``Unsupervised learning of finite mixture models,''
  \emph{IEEE Transactions on Pattern Analysis and Machine Intelligence},
  vol.~24, no.~3, pp. 381--396, 2002.

\bibitem{mclachlan2019}
G.~J. McLachlan, S.~X. Lee, and S.~I. Rathnayake, ``Finite mixture models,''
  \emph{Annual review of statistics and its application}, vol.~6, pp. 355--378,
  2019.

\bibitem{dempster1977}
A.~P. Dempster, N.~M. Laird, and D.~B. Rubin, ``Maximum likelihood from
  incomplete data via the em algorithm,'' \emph{Journal of the royal
  statistical society: series B (methodological)}, vol.~39, no.~1, pp. 1--22,
  1977.

\bibitem{yao2014}
W.~Yao, Y.~Wei, and C.~Yu, ``Robust mixture regression using the
  t-distribution,'' \emph{Computational Statistics \& Data Analysis}, vol.~71,
  pp. 116--127, 2014.

\bibitem{chamroukhi2016}
F.~Chamroukhi, ``Skew-normal mixture of experts,'' in \emph{2016 International
  Joint Conference on Neural Networks (IJCNN)}.\hskip 1em plus 0.5em minus
  0.4em\relax IEEE, 2016, pp. 3000--3007.

\bibitem{bai2012}
X.~Bai, W.~Yao, and J.~E. Boyer, ``Robust fitting of mixture regression
  models,'' \emph{Computational Statistics \& Data Analysis}, vol.~56, no.~7,
  pp. 2347--2359, 2012.

\bibitem{Naderi2023}
M.~Naderi, E.~Mirfarah, W.-L. Wang, and T.-I. Lin, ``Robust mixture regression
  modeling based on the normal mean-variance mixture distributions,''
  \emph{Computational Statistics \& Data Analysis}, vol. 180, p. 107661, 2023.

\bibitem{Li2018}
L.~Xiangrui and Z.~Dongxiao, ``Robust feature selection via l2, 1-norm in
  finite mixture of regression,'' \emph{Pattern Recognition Letters}, vol. 108,
  pp. 15--22, 2018.

\bibitem{yu2020selective}
C.~Yu, W.~Yao, and G.~Yang, ``A selective overview and comparison of robust
  mixture regression estimators,'' \emph{International Statistical Review},
  vol.~88, no.~1, pp. 176--202, 2020.

\bibitem{hathaway1993}
R.~J. Hathaway and J.~C. Bezdek, ``Switching regression models and fuzzy
  clustering,'' \emph{IEEE Transactions on fuzzy systems}, vol.~1, no.~3, pp.
  195--204, 1993.

\bibitem{krishnapuram1993}
R.~Krishnapuram and J.~M. Keller, ``A possibilistic approach to clustering,''
  \emph{IEEE transactions on fuzzy systems}, vol.~1, no.~2, pp. 98--110, 1993.

\bibitem{Zhang2004}
Z.~Jiangshe and L.~Yiu-Wing, ``Improved possibilistic c-means clustering
  algorithms,'' \emph{IEEE transactions on fuzzy systems}, vol.~12, pp.
  209--217, 2004.

\bibitem{kung2013}
C.-C. Kung, H.-C. Ku, and J.-Y. Su, ``Possibilistic c-regression models
  clustering algorithm,'' in \emph{2013 International Conference on System
  Science and Engineering (ICSSE)}.\hskip 1em plus 0.5em minus 0.4em\relax
  IEEE, 2013, pp. 297--302.

\bibitem{chang2016}
S.-T. Chang, K.-P. Lu, and M.-S. Yang, ``Stepwise possibilistic
  c-regressions,'' \emph{Information Sciences}, vol. 334, pp. 307--322, 2016.

\bibitem{zhao2018}
Y.~Zhao, P.-h. Wang, Y.-g. Li, and M.-y. Li, ``Fuzzy weighted c-harmonic
  regressions clustering algorithm,'' \emph{Soft Computing}, vol.~22, pp.
  4595--4611, 2018.

\bibitem{vidal2011subspace}
R.~Vidal, ``Subspace clustering,'' \emph{IEEE Signal Processing Magazine},
  vol.~28, no.~2, pp. 52--68, 2011.

\bibitem{cantzler1981}
H.~Cantzler, ``Random sample consensus (ransac),'' \emph{Institute for
  Perception, Action and Behaviour, Division of Informatics, University of
  Edinburgh}, vol.~3, 1981.

\bibitem{tsakiris2017ASC}
M.~C. Tsakiris and R.~Vidal, ``Algebraic clustering of affine subspaces,''
  \emph{IEEE transactions on pattern analysis and machine intelligence},
  vol.~40, no.~2, pp. 482--489, 2017.

\bibitem{agarwal2004}
P.~K. Agarwal and N.~H. Mustafa, ``K-means projective clustering,'' in
  \emph{Proceedings of the twenty-third ACM SIGMOD-SIGACT-SIGART symposium on
  Principles of database systems}, 2004, pp. 155--165.

\bibitem{tsakiris2015}
M.~C. Tsakiris and R.~Vidal, ``Dual principal component pursuit,'' in
  \emph{Proceedings of the IEEE International Conference on Computer Vision
  Workshops}, 2015, pp. 10--18.

\bibitem{bezdek1981}
J.~C. Bezdek, C.~Coray, R.~Gunderson, and J.~Watson, ``Detection and
  characterization of cluster substructure ii. fuzzy c-varieties and convex
  combinations thereof,'' \emph{SIAM Journal on Applied Mathematics}, vol.~40,
  no.~2, pp. 358--372, 1981.

\bibitem{Chen2020}
L.~Guo, L.~Chen, X.~Lu, and C.~L.~P. Chen, ``Membership affinity lasso for
  fuzzy clustering,'' \emph{IEEE Transactions on Fuzzy Systems}, vol.~28,
  no.~2, pp. 294--307, Feb 2020.

\bibitem{danon2005}
L.~Danon, A.~Diaz-Guilera, J.~Duch, and A.~Arenas, ``Comparing community
  structure identification,'' \emph{Journal of statistical mechanics: Theory
  and experiment}, vol. 2005, no.~09, p. P09008, 2005.

\bibitem{steinley2004}
D.~Steinley, ``Properties of the hubert-arable adjusted rand index.''
  \emph{Psychological methods}, vol.~9, no.~3, p. 386, 2004.

\bibitem{amigo2009}
E.~Amig{\'o}, J.~Gonzalo, J.~Artiles, and F.~Verdejo, ``A comparison of
  extrinsic clustering evaluation metrics based on formal constraints,''
  \emph{Information retrieval}, vol.~12, pp. 461--486, 2009.

\bibitem{cohen1984}
E.~A. Cohen, ``Some effects of inharmonic partials on interval perception,''
  \emph{Music Perception}, vol.~1, no.~3, pp. 323--349, 1984.

\bibitem{campbell1974}
N.~Campbell and R.~Mahon, ``A multivariate study of variation in two species of
  rock crab of the genus leptograpsus,'' \emph{Australian Journal of Zoology},
  vol.~22, no.~3, pp. 417--425, 1974.

\bibitem{wold1987PCA}
S.~Wold, K.~Esbensen, and P.~Geladi, ``Principal component analysis,''
  \emph{Chemometrics and Intelligent Laboratory Systems}, vol.~2, no.~1, pp.
  37--52, 1987.

\end{thebibliography}

\end{document}